\documentclass[letterpaper]{article}
\usepackage{uai2019}
\usepackage[margin=1in]{geometry}

\usepackage{epsf}
\usepackage{fancyhdr}
\usepackage{graphics}
\usepackage{graphicx}
\usepackage{psfrag}
\usepackage{microtype}
\usepackage{subfigure}
\usepackage{algorithmic}
\usepackage{color,xcolor}
\usepackage[linesnumbered,ruled]{algorithm2e}
\usepackage{color}

\usepackage{wrapfig}
\usepackage{float}

\usepackage{amsthm}
\usepackage{amsfonts}
\usepackage{amsmath}
\usepackage{amssymb,bbm}
\usepackage{multirow}
\usepackage{balance}

\usepackage{url}
\usepackage{hyperref}
\hypersetup{
    colorlinks=true,
    linkcolor=blue,
    filecolor=blue,      
    urlcolor=blue,
}

\usepackage{nicefrac}
\usepackage{comment}

 \usepackage{tabularx}%

\usepackage{enumitem}
\usepackage{booktabs}
\usepackage{caption}

\usepackage{bm,bbm}
\usepackage{mathtools}

\newcommand{\Rspace}{\mathbb{R}}
\newtheorem{assumption}{Assumption}
\newtheorem{proposition}{Proposition}
\newtheorem{definition}{Definition}

\newtheorem{remark}{Remark}

\def\argmin{\textnormal{arg} \min}

\newcommand{\norm}[1]{\left\lVert#1\right\rVert}
\renewcommand{\algorithmicrequire}{\textbf{Input:}}
\renewcommand{\algorithmicensure}{\textbf{Output:}}

\def\TM{\texttt{T}}

\def\TW{\textnormal{TW}}

\def\RR{\mathbb{R}}

\def\Pp{\mathcal{P}}
\def\Ss{\mathcal{S}}

\def\Tt{\mathcal{T}}

\def\yb{\mathbf{y}}
\def\ub{\mathbf{u}}
\def\betab{\bm{\beta}}
\def\Ncal{\mathcal{N}}
\def\zero{\mathbf{0}}
\def\diag{\textnormal{diag}}

\title{Tree-Wasserstein Barycenter \\ for Large-Scale Multilevel Clustering and Scalable Bayes}

\author{Tam Le$^{\star}$ \\ RIKEN AIP \And Viet Huynh$^{\star}$ \\ Monash University \And Nhat Ho$^{\star}$ \\ UC Berkeley \And Dinh Phung \\ Monash University \And Makoto Yamada \\ RIKEN AIP \\ Kyoto University} 

\begin{document}

\maketitle




\begin{abstract}
We study in this paper a variant of Wasserstein barycenter problem, which we refer to as tree-Wasserstein barycenter, by leveraging a specific class of ground metrics, namely tree metrics, for Wasserstein distance. Drawing on the tree structure, we propose an efficient algorithmic approach to solve the tree-Wasserstein barycenter and its variants. The proposed approach is not only fast for computation but also efficient for memory usage. Exploiting the tree-Wasserstein barycenter and its variants, we scale up multi-level clustering and scalable Bayes, especially for large-scale applications where the number of supports in probability measures is large. Empirically, we test our proposed approach against other baselines on large-scale synthetic and real datasets.
\end{abstract}

\let\thefootnote\relax\footnotetext{$^\star$: equal contribution.}



\section{Introduction}
In the big data era, large-scale datasets have become a norm in several applications in statistics and machine learning. Recently, optimal transport (OT) distance has been employed as a popular and powerful tool in these applications, including computer graphics~\cite{Solomon_2015_Convolutional, solomon2019optimal}, deep learning~\cite{Arjovsky-2017-Wasserstein, Courty-2017-Optimal,  Tolstikhin-2018-Wasserstein}, and computational biology \cite{schiebinger2019optimal}.

In principle, OT distance between two discrete probability measures can be formulated as a linear programming problem, which can be solved by interior point methods. However, as pointed out in several works~\cite{Lee-2014-Path, Pele-2009-Fast}, these methods are not scalable when the number of supports of input probability measures is large. Recently, Cuturi~\cite{Cuturi-2013-Sinkhorn} proposed an entropic regularization of OT, which we refer to as entropic OT, as an efficient way to solve the scalability issue of OT. Given the special structure of its dual form, the entropic OT can be solved by the celebrated Sinkhorn algorithm~\cite{Sinkhorn-1974-Diagonal}. Due to the favorable practical performance of the Sinkhorn algorithm, several works have analyzed its computational complexity~\cite{Altschuler-2017-Near, Dvurechensky-2018-Computational}, and further improve its performance~\cite{altschuler2019massively, Lin_2019_acceleration}.

Another direction for scaling up the computation of OT distance includes sliced Wasserstein (SW) distance~\cite{Rabin_2019_sliced}. The idea of SW distance is to project supports of probability measures into one-dimensional space and then using the closed-form expression of univariate Wasserstein distance as a variant for OT distance. Given its fast computation, SW distance has been employed to numerous problems in deep generative models~\cite{Deshpande_2018_generative, Kolouri_2019_sliced, Wu_2019_sliced}. However, due to the one-dimensional projection, SW distance is limited to retain the high-dimensional structure of support data distributions~\cite{Bonneel_2015_sliced, pmlr-v97-liutkus19a}.  

Recently, Le et al.~\cite{le2019tree} proposed tree-(sliced)-Wasserstein distance in which the SW distance is a particular instance. The idea of tree-Wasserstein (TW) distance is to use a specific class of ground metrics, namely tree metrics, for OT distance which yields a closed-form solution. Consequently, the TW distance also enjoys a fast computation as that of univariate OT. In addition, a tree metric is constructed on the original space of supports. For example, one can use the clustering-based tree metric sampling~\cite{le2019tree} (\S4) which directly leverages a distribution of supports to construct a tree metric. Therefore, TW distance preserves the structure of the probability measures better than that of SW distance, since choosing a tree has far more degrees of freedom than choosing a line, especially in high-dimensional support data spaces. 

Going beyond OT distance, Wasserstein barycenter, a problem of finding optimal probability measure to minimize its OT distances to a given set of probability measures, becomes of interest for several applications, e.g., multilevel clustering~\cite{Ho-ICML-2017}, and scalable Bayes inference~\cite{Srivastava-2018-Scalable}. Even though several algorithms were proposed to compute Wasserstein barycenter as well as its corresponding entropic regularized version~\cite{Benamou_2015, Cuturi-2014-Fast, Kroshnin_2019_complexity}, large-scale applications of these algorithms have still been challenging, especially when supports are in high-dimensional spaces.

In this work, we follow the research direction of \cite{le2019tree}, using a specific class of ground metrics, namely tree metrics, for OT to address the scalability issue of computational Wasserstein barycenter. We refer to our approach as the tree-Wasserstein barycenter. Our contribution is two-fold: (i) by leveraging tree structure, we propose efficient algorithmic approaches for tree-Wasserstein barycenter and its variants, (ii) and we apply our proposed algorithms to scale up multilevel clustering and scalable Bayes for large-scale applications where the number of supports is large.

The remainder of the paper is organized as follows. In Section~\ref{sec:TW}, we give a brief review of tree metrics and TW distance. We then present algorithms for solving TW barycenter and its variants in Section~\ref{sec:TW_barycenter}. After that, we apply the proposed algorithms for large-scale multilevel clustering and scalable Bayes in Section~\ref{sec:TW_applications}. We next show several experiment results in Section~\ref{sec:experiments} before having a conclusion in Section~\ref{sec:discussion}.

\textbf{Notation.} We denote $[n] = \{1, 2, \ldots, n \}$ for any $n \in \mathbb{N}_{+}$. For any discrete probability distribution $G$, the notation $|G|$ stands for the number of supports of $G$. For any $x \in \RR^{d}$, $\norm{x}_1$ is the $\ell_1$-norm of $x$.


\section{Tree Wasserstein (TW) distance}
\label{sec:TW}
In this section, we give a brief review of tree metric~\cite{semple2003phylogenetics}, and tree-Wasserstein distance~\cite{le2019tree}.


\subsection{Tree metric}

\begin{definition}
Given a finite set $\Omega$, a metric $d:\Omega\times\Omega\rightarrow \RR_{+}$ is a tree metric on $\Omega$ if there exists tree $\Tt$ with non-negative edge lengths such that $\forall x \in \Omega$, $x$ is a node in $\Tt$, and $\forall x, z \in \Omega$, $d(x, z)$ equals to the length of the (unique) path between $x$ and $z$ in tree $\Tt$.
\end{definition}

\begin{figure}
  \begin{center}
    \includegraphics[width=0.21\textwidth]{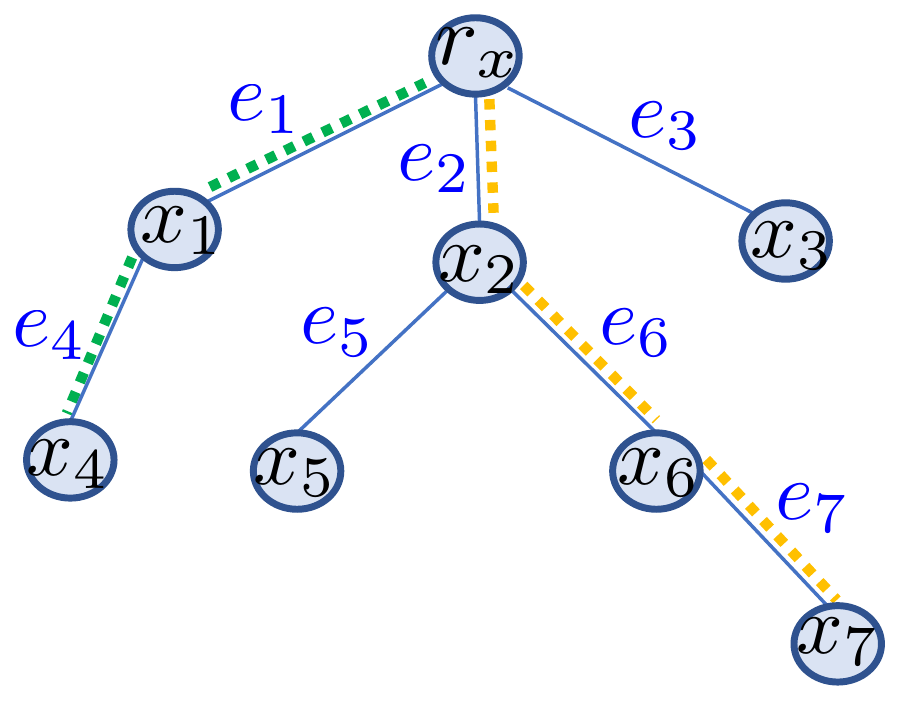}
  \end{center}
  \caption{An illustration for tree metric space. Root $r_x$ is at depth level $1$, while $x_4, x_5, x_6$ are at depth level $3$. $\Pp(r_x, x_4)$ contains $e_1, e_4$ (the green dot path). $\Gamma(x_2) = \left\{x_2, x_5, x_6, x_7 \right\}$. For an edge $e_6$, $v_{e_6} = x_6$, and $u_{e_6} = x_2$. For node $x_2$, there are 3 directions which one can proceed to leave $x_2$, i.e., directions from $x_2$ towards $r_x$, from $x_2$ towards $x_5$, and from $x_2$ towards $x_6$.}
  \label{fg:TreeMetric}
\end{figure}

Let $d_{\TM}$ be the tree metric on tree $\Tt$, rooting at node $r$. Given a node $x \in \Tt$, we denote $\Gamma(x)$ as a set of nodes for a subtree of $\Tt$ rooted at $x$, i.e., $\Gamma(x) = \left\{z \in \Tt \mid x \in \Pp(r, z) \right\}$ where $\Pp(r, z)$ is the (unique) path between root $r$ and node $z$ in $\Tt$. For an edge $e$ of tree $\Tt$, let $v_e$ be the deeper-level node, $u_e$ is the other node, and $w_e=d_{\TM}(u_e, v_e)$ be the non-negative length of $e$, as illustrated in Figure~\ref{fg:TreeMetric}.

\subsection{Tree Wasserstein distance}\label{sec:tree_Wasserstein}

Let $\mu, \nu$ be two Borel probability measures on $\Tt$, the \emph{tree Wasserstein distance} $d_{\TW}$ between $\mu$ and $\nu$ takes the form
\begin{align}
\hspace{-1 em}    d_{\TW}(\mu, \nu; \Tt) : = \inf \limits_{\pi \in \Pi( \mu, \nu)} \int_{\Tt \times \Tt} d_{\TM}(x, y) \pi(\text{d}x, \text{d}y),
\end{align}
where $\Pi( \mu, \nu)$ is the set of probability distributions $\pi$ on the product space $\Tt \times \Tt$ such that its marginal distributions are respectively $\mu$ and $\nu$. As shown in~\cite{le2019tree}, the tree Wasserstein (TW) distance admits the closed-form solution as follows:
\begin{equation}
    \label{equ:TW}
    d_{\TW}(\mu, \nu; \Tt) = \sum_{e \in \Tt} w_e \left|\mu(\Gamma(v_e)) - \nu(\Gamma(v_e)) \right|.
\end{equation}
When the context of tree metric space is clear, we drop $\Tt$ and write $d_{\TW}(\mu, \nu)$ for $d_{\TW}(\mu, \nu; \Tt)$.

\section{TW barycenter and its variants}\label{sec:TW_barycenter}

In this section, we develop efficient algorithms for solving the TW barycenter and its variant problems.  

\subsection{Tree-Wasserstein barycenter}\label{sec:unconstrained_tree_barycenter}
Given probability measures $\mu_{i} \mid_{i \in [n]}$ whose supports are in a tree metric space $(\Tt, d_{\TM})$ and their corresponding weights $p_{i} \mid_{i \in [n]}$ such that $p_{i} \geq 0 \mid_{i \in [n]}$ and $\sum_{i = 1}^{n} p_{i} = 1$, the TW barycenter $\bar{\mu}$ is formulated as follows: 
\begin{equation}
\label{equ:TW_Barycenter}
\bar{\mu} \in \argmin_{\mu} \biggr(\sum_{i = 1}^{n} p_i d_{\TW}(\mu, \mu_i; \Tt)\biggr),
\end{equation}
where the minimum is taken with respect to any probability measure in $(\Tt, d_{\TM})$.

Plugging Equation \eqref{equ:TW} in to Equation~\eqref{equ:TW_Barycenter}, we have
\small{\begin{equation}
\label{equ:equiv_TW_Barycenter}
 \bar{\mu} \in \argmin_{\mu} \biggr(\! \sum_{i = 1}^{n} p_i \sum_{e \in \Tt} w_e \left|\mu(\Gamma(v_e)) - \mu_{i}(\Gamma(v_e)) \right| \!\biggr).
\end{equation}}

For any probability measure $\mu$ in $(\Tt, d_{\TM})$, we observe that $w_e \mu(\Gamma(v_e)) \mid_{e \in \Tt}$ can be regarded as a tree mapping for $\mu$ into $\RR^{m}$ where $m$ is the number of edges in tree $\Tt$. Then, TW distance $d_{\TW}$ between two probability measures is equivalent to the $\ell_1$ distance between their corresponding mappings in $\RR^{m}$. The following result shows that one can retrieve $\mu$ for its corresponding tree mapping.

\begin{proposition}
\label{prop:inverse_map}
Given a tree mapping $\alpha \in \RR^{m}$ of some probability measure $\mu$ in tree $\Tt$ with $m$ edges, i.e., each dimension $\alpha_{e} = w_e \mu(\Gamma(v_e))$ for each edge $e$ in $\Tt$, then one can recover $\mu$, as follow: 

Let $x_i$ be a node in $\Tt$, the corresponding weight $a_i$ for $\delta_{x_i}$ of $\mu=\sum_i a_i \delta_{x_i}$ takes the form:
\begin{equation}\label{equ:weight_recover}
a_i = \sum_{e \in \Tt, v_e = x_i} {\alpha_{e}}/{w_e} - \sum_{e \in \Tt, u_e = x_i} {\alpha_{e}}/{w_e}.
\end{equation}
\end{proposition}
\begin{proof}
From \cite{le2019tree}, the total ``mass" flowing through edge $e$ is equal to total ``mass" in subtree $\Gamma(v_e)$. For a node $x$ in $\Tt$, let $\Ss_{x}$ be the set of children nodes of $x$, then 
\[
a_x = \mu(\Gamma(x)) - \sum_{\tilde{x} \in \Ss_{x}} \mu(\Gamma(\tilde{x})).
\]
Additionally, we also have $\alpha_e = w_e \mu(\Gamma(v_e))$. Hence, one can recover weight $a_i$ for each $\delta_{x_i}$ in $\mu$ as in Equation~\eqref{equ:weight_recover}.
\end{proof}

Let $h: \mu \mapsto w_e \mu(\Gamma(v_e)) \mid_{e \in \Tt}$, and denote $\bar{z} = h(\bar{\mu}), z = h(\mu), z_i = h(\mu_i) \mid_{i \in [n]} \in \RR^m$. Equation~\eqref{equ:equiv_TW_Barycenter} can be reformulated as the following barycenter problem for $\ell_1$ distance in $\RR^{m}$:
\begin{equation}
\label{equ:L1_Barycenter}
\bar{z} \in \argmin_{z \in \RR^{m}} \biggr( \sum_{i = 1}^{n} p_i \norm{z - z_i}_1 \biggr).
\end{equation}
Since the $\ell_1$ distance is separable with respect to dimensions of data points, Equation~\eqref{equ:L1_Barycenter} can be solved separately for each dimension. More specifically, for each dimension corresponding to an edge $e$ in $\Tt$, the optimal component $\bar{z}_{e}$ is the weighted geometric median (WGM) of $z_{ie} \mid_{i \in [n]}$ with the corresponding weights $p_{i} \mid_{i \in [n]}$:
\begin{align}\label{equ:L1_Barycenter_dim}
    \bar{z}_{e} \in \argmin_{z_{e} \in \RR} \biggr( \sum_{i = 1}^{n} p_{i}|z_{e} - z_{ie}|\biggr).
\end{align}

Moreover, $\bar{\mu}$ is a probability measure on tree $\Tt$ as in Equation~\eqref{equ:TW_Barycenter}, and recall that $\bar{z}$ is the corresponding tree mapping of TW barycenter $\bar{\mu}$ on tree $\Tt$ via the mapping $h$. Therefore, we can retrieve the probability measure $\bar{\mu}$ from $\bar{z}$ by applying Proposition~\ref{prop:inverse_map} which is summarized in Algorithm~{\ref{alg:InverseMapping}}. Additionally, in practice, we use Algorithm~\ref{alg:WGM} to solve Equation \eqref{equ:L1_Barycenter_dim} which outputs the minimum value when the weighted geometric median is not unique. Finally, the pseudo-code for TW barycenter is given in Algorithm~{\ref{alg:TW_Barycenter}}.

\begin{algorithm}[t] 
\caption{Retrieve measure from tree mapping} 
\label{alg:InverseMapping} 
\begin{algorithmic}[1] 
    \REQUIRE A feature map $z$ where each dimension $z_{(e)}$ is corresponding the edge $e$ in tree $\Tt$ rooted at $r$.
    \ENSURE Probability measure $\mu$ on tree $\Tt$.
    
    \STATE For root $r$: $a_r \leftarrow 1 - \sum_{e \in \Tt, u_e = r} {z_{(e)}}/{w_e}$.
    \STATE Set $\mu \leftarrow a_r \delta_{r}$.
    \FOR{\textbf{each} node $x$ in $\Tt$}
	\STATE $a_x \leftarrow \sum_{e \in \Tt, v_e = x} {z_{(e)}}/{w_e} - \sum_{e \in \Tt, u_e = x} {z_{(e)}}/{w_e}$.
	\STATE $\mu \leftarrow \mu + a_x \delta_{x}$. 
    \ENDFOR
\end{algorithmic}
\end{algorithm}

\begin{algorithm}[t] 
\caption{Tree-Wasserstein Barycenter} 
\label{alg:TW_Barycenter} 
\begin{algorithmic}[1] 
    \REQUIRE Probability measures $\mu_i \mid_{1 \le i \le n}$, weights $p_i \mid_{1 \le i \le n}$, and tree $\Tt$.
    \ENSURE TW barycenter $\bar{\mu}$
    
    \FOR{\textbf{each} $\mu_i$}
	\STATE Compute tree mapping $z_i \leftarrow h(\mu_i)$. 
    \ENDFOR
    \STATE Compute component-wise WGM $\bar{z}$ for $z_i \mid_{i \in [n]}$ using Algorithm~\ref{alg:WGM} for each dimension.
    \STATE Recover TW barycenter $\bar{\mu}$ from $\bar{z}$ using Algorithm~\ref{alg:InverseMapping}.
	
\end{algorithmic}
\end{algorithm}

\begin{algorithm}[t] 
\caption{Weighted geometric median (WGM)} 
\label{alg:WGM} 
\begin{algorithmic}[1] 
    \REQUIRE Scalar values $a_i \mid_{1 \le i \le n}$, and their corresponding weights $p_i \mid_{1 \le i \le n}$ where $p_i \ge 0$ and $\sum_{i=1}^{n} p_i = 1$.
    \ENSURE Weighted geometric median $\bar{a}$.
    
    \STATE Let $\sigma$ be the permutation which sorts $a_i \mid_{1 \le i \le n}$ in ascending order.
    \IF{$p_{\sigma(1)} \ge 1/2$}
    	\STATE $\bar{a} \leftarrow a_{\sigma(1)}$.
    \ELSIF{$p_{\sigma(n)} \ge 1/2$}
    	\STATE $\bar{a} \leftarrow a_{\sigma(n)}$.
    \ELSE
    	\STATE Accumulate $\hat{p} \leftarrow p_{\sigma(1)}$, set $t \leftarrow 2$, $flag \leftarrow true$.
	\WHILE{$flag$}  
		\IF{$1 - \hat{p} - p_{\sigma(t)} \le 1/2$}
			\STATE $\bar{a} \leftarrow a_{\sigma(t)}$, $flag \leftarrow false$.
		\ELSE
			\STATE $\hat{p} \leftarrow \hat{p} + p_{\sigma(t)}$, $t \leftarrow t + 1$.
		\ENDIF 
	\ENDWHILE
    \ENDIF
	
\end{algorithmic}
\end{algorithm}

\subsection{Tree-Wasserstein barycenter with a constraint on the number of supports}
\label{sec:constrain_tree_bary}


In this section, we consider a variant of TW barycenter problem where the number of supports of the barycenter is bounded by some given constant. This variant problem appears in several applications, e.g., multilevel clustering problem with images and documents (see Section~\ref{sec:tree_wasserstein_multilevel} for the details). 

Given probability measures $\mu_{i} \mid_{i \in [n]}$ in $(\Tt, d_{\TM})$ with corresponding weights $p_{i} \mid_{i \in [n]}$, the TW barycenter $\bar{\mu}_{c}$ whose number of supports are bounded by $\kappa$ is formulated as follows:
\begin{equation}
\label{equ:cons_TW_Barycenter}
\bar{\mu}_{c} \in \argmin_{|\mu| \leq \kappa} \sum_{i = 1}^{n} p_i d_{\TW}(\mu, \mu_i),
\end{equation}
where $\kappa \geq 1$ is a pre-defined positive integer parameter and recall that $|\mu|$ is the number of supports for measure $\mu$. We relax the problem in Equation~\eqref{equ:cons_TW_Barycenter} into
\begin{equation}
\label{equ:cons_TW_Barycenter}
\hspace{-1 em} \bar{\mu}_{c} \in \argmin_{|\mu| \leq \kappa} d_{\TW}\!\left(\mu, \argmin_{\bar{\mu}} \sum_{i = 1}^{n} p_i d_{\TW}(\bar{\mu}, \mu_i)\right).
\end{equation}
Intuitively, we would like to leverage the efficient computation of TW barycenter $\bar{\mu} = \sum_{j \in [m]} b_j \delta_{z_j}$ in Equation~\eqref{equ:TW_Barycenter}, and then find a probability measure whose number of supports are bounded by $\kappa$ closest to $\bar{\mu}$ in $(\Tt, d_{\TM})$. If $m \le \kappa$, then $\bar{\mu}_c = \bar{\mu}$, otherwise we can find $\bar{\mu} = \sum_{i \in [\kappa] a_i \delta_{x_i}}$ as a solution of $\kappa$-means for supports $z_j \mid_{j \in [m]}$ and corresponding weights $b_j \mid_{j \in [m]}$ w.r.t. tree metric $d_{\TM}$, defined as follow: 

\begin{equation}
    \argmin_{S} \sum_{i \in [\kappa]} \sum_{z_j \in S_i} b_j d_{\TM}^2(x_i, z_j), 
\end{equation}
where $S = \left\{S_i \mid_{i \in [\kappa]} \right\}$, and $x_i$ is the weighted mean w.r.t. $d_{\TM}$ of the set $S_i$ for all $i \in [\kappa]$. Additionally, we can obtain $a_i \mid_{i \in [\kappa]}$ in $\bar{\mu}$ by 
\[
a_i = \sum_{j \mid z_j \in S_i} b_j.
\]

\paragraph{Center of mass of a measure on a tree.} Computing a weighted mean $x$ w.r.t. $d_{\TM}$ for a set of supports $S_i$ and their corresponding weights $\{b_j \mid_{j \mid z_j \in S_i}\}$ is equivalent to find a \textit{center of mass} of a measure on tree $\Tt$.

Given a node $u \in \Tt$, $\sigma$ is an associated direction which one can proceed to leave $u$. For a measure $\nu = \sum b_j \delta_{z_j}$, a node $v \in \Tt$ such that $v \in \Tt$, $v \in \nu$, $v \neq u$, then $v$ is an element of a set $R_{\nu}(u, \sigma)$ if there is a unique path connecting $u$ and $v$ departing from $u$ in the direction $\sigma$, otherwise $v$ is an element of a set $U_{\nu}(u, \sigma)$. We introduce 
\begin{align}
    \Delta_{\nu}(u, \sigma) &:= \sum_{j \mid z_j \in U_{\nu}(u, \sigma)} b_j d_{\TM}(u, z_j) \nonumber \\ & \qquad \qquad - \sum_{j \mid z_j \in R_{\nu}(u, \sigma)} b_j d_{\TM}(u, z_j)
\end{align}

Following \cite{evans2012phylogenetic}, if there exists a node $u \in \Tt$ such that 
\[
\Delta_{\nu}(u, \sigma) \ge 0
\]
for all directions $\sigma$ associated with node $u$, then $x = u$ is the center of mass of $\nu$, otherwise, there exists a unique edge $e \in \Tt$ such that 
\[
\Delta_{\nu}(u_e, \alpha) < 0, \qquad \Delta_{\nu}(v_e, \beta) < 0
\]
where $\alpha$ is the direction from $u_e$ towards $v_e$, and $\beta$ is the direction from $v_e$ towards $u_e$, and the center of mass $u$ of $\nu$ is on the edge $e$ of $\Tt$ such that 
\[
d_{\TM}(u_e, u) = -\Delta_{\nu}(u_e, \alpha).
\]
For simplicity, we set the center of mass $x$ of $\nu$ as: $x=u_e$ if $d_{\TM}(u_e, u) < d_{\TM}(v_e, u)$, otherwise $x=v_e$, note that 
\[
d_{\TM}(v_e, u) = d_{\TM}(u_e, v_e) - d_{\TM}(u_e, u).
\]

\begin{algorithm}[t] 
\caption{TW barycenter with a constraint on the number of supports} 
\label{alg:Constrained_TW_Barycenter} 
\begin{algorithmic}[1] 
    \REQUIRE Probability measures $\mu_i \mid_{1 \le i \le n}$, their corresponding weights $p_i \mid_{1 \le i \le n}$ where $p_i \ge 0$ and $\sum_{i=1}^{n} p_i = 1$, $\kappa$ is the constrained number of supports of the barycenter.
    \ENSURE TW barycenter $\bar{\mu}_c$ with at most $\kappa$ supports
    
    \STATE Using Algorithm~\ref{alg:TW_Barycenter} to compute TW barycenter $\bar{\mu}$.
    
    \IF{$\left|\bar{\mu}\right| \le \kappa$}
        \STATE $\bar{\mu}_c \leftarrow \bar{\mu}$.
    \ELSE
        \STATE Find $\bar{\mu}_c$ by applying $\kappa$-means clustering w.r.t. tree metric $d_{\TM}$ for supports of $\bar{\mu}$ and their corresponding weights where supports and weights of $\bar{\mu}_c$ are centroids, and their corresponding masses of clusters respectively.
    \ENDIF
\end{algorithmic}
\end{algorithm}

\subsection{Multiple-tree variants of TW barycenter by sampling tree metrics}\label{sec:multiple_tree_TW_barycenter}

In this section, we further derive multiple-tree variants for TW barycenter by averaging the TW barycenter obtained on each tree metric. In particular, we apply the clustering-based tree metric method~\cite{le2019tree} to randomly sample tree metric $\Tt_i \mid_{i \in [k]}$, let $\hat{\nu}_{i} \mid_{i \in [k]}$ be TW barycenter (Equation~\eqref{equ:TW_Barycenter} or Equation~\eqref{equ:cons_TW_Barycenter}) in each corresponding tree metric space $\Tt_i \mid_{i \in [k]}$, the corresponding multiple-tree variant $\hat{\mu}$ of TW barycenter is defined as follow:
\begin{equation}
\hat{\mu} = \frac{1}{k} \sum_{i \in [k]} \hat{\nu}_i.
\end{equation}

Similar to tree-sliced-Wasserstein~\cite{le2019tree}, the multiple-tree variants of TW barycenter can reduce the clustering sensitivity effect for the clustering-based tree metric sampling due to averaging over many TW barycenter on corresponding random tree metric space.

For tree metric sampling in multiple-tree variants of TW barycenter, the clustering-based tree metric sampling~\cite{le2019tree} is very fast in practise, and its computation is negligible. As discussed in \cite{le2019tree}, one can use any clustering methods for the clustering-based tree metric sampling. We followed Le et al.~\cite{le2019tree} to use the suggested farthest-point clustering due to its fast computation (e.g., its complexity is $O(m\log k)$ to cluster $m$ points into $k$ clusters). Therefore, the complexity of the clustering-based tree metric sampling is $O(m H_{\Tt} \log k)$ when one uses the same $k$ for the farthest-point clustering, $H_{\Tt}$ as the predefined deepest level of tree $\Tt$ for $m$ input data points.

\begin{remark}
We emphasize that the multiple-tree variants of TW barycenter is defined by averaging corresponding TW barycenters on each random tree metric, and it is different to the tree-sliced-Wasserstein (TSW) barycenter where the averaging is on corresponding TW distances. In addition, for the TW barycenters (Equation~\eqref{equ:TW_Barycenter}), in case those random tree metrics are independent, e.g. nodes in the corresponding trees (finite set of nodes) of those tree metrics are disjoint to each other, then the proposed multiple-tree variant is equivalent to the TSW barycenter. For the TW barycenters with a constraint on the number of supports (Equation~\eqref{equ:cons_TW_Barycenter}), in case one use the averaging over random tree metrics for the subgradient computation, then one will have TSW barycenter version.
\end{remark}

\section{Large-scale multilevel clustering and scalable Bayes with tree-Wasserstein barycenter and its variants}\label{sec:TW_applications}

\subsection{Multilevel clustering}
\label{sec:tree_wasserstein_multilevel}

We first discuss an application of TW barycenter and its variants to the multilevel clustering problem, which arises in various real applications in text data and computer vision. In particular, we consider $m$ groups of data points $X_{ij}$ where $i \in [m]$ (group index), $j \in [n_{i}]$ (data point index in a group), and $n_i$ is the number of data points in group $i$. The main goal of the multilevel clustering problem is to simultaneously partition data $X_{ij}$ in each group $i$ and cluster the groups. There are two popular approaches to solving the multilevel clustering problem. The first approach is through Bayesian hierarchical models, such as hierarchical Dirichlet process~\cite{huynh2016scalable, Teh-et-al06}, nested Dirichlet process~\cite{Abel-2008}, or a combination of these processes~\cite{huynh2016scalable, Vu-2014, Wulsin-2016}. The second approach is through an optimization approach based on optimal transport (OT) distances~\cite{Ho-ICML-2017}, which is amendable to large-scale settings. In this paper, we employ the second approach with TW distance and demonstrate that our model has a much faster running time than previous models in the same vein.

To ease the ensuing presentation, we denote 
\[
P_{i} : = \frac{1}{n_{i}} \sum_{j = 1}^{n_{i}} \delta_{X_{ij}}
\]
as an empirical measure associated with group $i$. We assume that there are at most $k_{i}$ clusters in each group $i$ for $i \in [m]$ while there are at most $K$ clusters for the $m$ groups where $K \geq 2$. The idea of our model is similar in spirit to that of multilevel Wasserstein means (MWM) in~\cite{Ho-ICML-2017}. In each group $i \in [m]$, we partition the data $X_{ij}$ based on TW distance, namely, we seek for a discrete probability measure $G_{i}$ with at most $k_{i}$ supports such that it minimizes the TW distance: 
\[
d_{\TW}(G_{i}, P_{i}).
\]
That step is referred to as \textit{local clustering}. By viewing $G_{i}$ as points in the probability space of discrete probability measures, we can cluster these groups utilizing K-means using the TW distance. More precisely, we determine a set of discrete probability distributions $\mathcal{Q} = \left\{Q_{k} \mid_{k \in [K]}\right\}$ that minimizes the objective function:
\[
\frac{1}{m} \sum_{i = 1}^{m} \min_{k \in [K]} d_{\TW}(Q_{k}, G_{i}).
\]
This step is referred to as \textit{global clustering}. To capture the sharing information among groups in global clustering, we will jointly optimize the probability measures $G_{i}$ from local clustering and the set of probability distributions $\mathcal{Q}=\{Q_{k} \mid_{k \in [K]}\}$ from global clustering. More precisely, we derive the following objective function
\begin{align}
    \inf_{G_{i}, \mathcal{Q}} \sum_{i = 1}^{m} d_{\TW}(G_{i}, P_{i}) + \frac{\lambda}{m} \sum_{i = 1}^{m} \min_{k \in [K]} d_{\TW}(Q_{k}, G_{i}), \label{eq:tree_multilevel}
\end{align}
where $\lambda > 0$ is used to balance the losses from the local clustering and global clustering. The infimum in Equation~\eqref{eq:tree_multilevel} is taken with respect to discrete probability measures $G_{i}$ with at most $k_{i}$ supports. We call the above objective function \textit{tree-Wasserstein multilevel clustering}. 

To solve the TW multilevel clustering in~\eqref{eq:tree_multilevel}, we employ an alternating optimization approach, namely, we fix $\mathcal{Q}$ and optimize $G_{i} \mid_{i \in [m]}$ using the TW barycenter with a constraint on the number of supports (Section~\ref{sec:constrain_tree_bary}), and then we fix $G_{i} \mid_{i \in [m]}$ and optimize $\mathcal{Q}$ using the TW barycenter (Section~\ref{sec:unconstrained_tree_barycenter}). Furthermore, one can use multiple-tree variants of TW barycenter (Section~\ref{sec:multiple_tree_TW_barycenter}) by sampling several random tree metrics to reduce the cluster sensitivity effect. The pseudo-code of our algorithm is presented in Algorithm~\ref{alg:tree_Wasserstein_multilevel}. 

\begin{algorithm}[tb]
   \caption{Tree-Wasserstein multilevel clustering}
   \label{alg:tree_Wasserstein_multilevel}
\begin{algorithmic}
   \STATE {\bfseries Input:} Data $X_{ij}$, the number of clusters $k_{i}, K$.
   \STATE {\bfseries Output:} Local probability measures $G_{i}$ and set of global probability measures $\mathcal{Q}$.
   \STATE Initialize measures $G_{i}^{(0)}$, elements $Q_{k}^{(0)}$ of $\mathcal{Q}^{(0)}$, $t \leftarrow 0$.
   \WHILE{$G_{i}^{(t)}, \mathcal{Q}^{(t)}$ have not converged}
   \STATE \textit{\% Update $G_{i}^{(t)} \mid_{i \in [m]}$:}
   \STATE $j_{i} \leftarrow \mathop {\arg \min}\limits_{k \in [K]}{d_{\TW}(G_{i}^{(t)},Q_{k}^{(t)})}$.
   \begin{align*}
   G_{i}^{(t+1)} \leftarrow 
   & \mathop {\arg \min }\limits_{|G_{i}| \leq k_{i}}{d_{\TW}(G_{i},P_{i})} +  \frac{\lambda}{m} d_{\TW} (G_{i}, Q_{j_{i}}^{(t)}).
   \end{align*}
   
   \STATE \textit{\% Update $Q_{k}^{(t)} \mid_{k \in [K]}$:}
   \STATE $j_{i} \leftarrow \mathop {\arg \min}\limits_{k \in [K]}{d_{\TW}(G_{i}^{(t+1)}, Q_{k}^{(t)})}$ for $i \in [m]$.
   \STATE $C_{k} \leftarrow \left\{\ell: j_{\ell}= k \right\}$ for $k \in [K]$.
   \STATE $Q_{k}^{(t+1)} \leftarrow \mathop {\arg \min }\limits_{Q_{k}}{\sum \limits_{\ell \in C_{k}}{d_{\TW}(Q_{k}, G_{\ell}^{(t+1)})}}$.
   \STATE $t \leftarrow t+1$.
   \ENDWHILE
\end{algorithmic}
\end{algorithm}


\begin{figure*}
  \begin{center}
    \includegraphics[width=\textwidth]{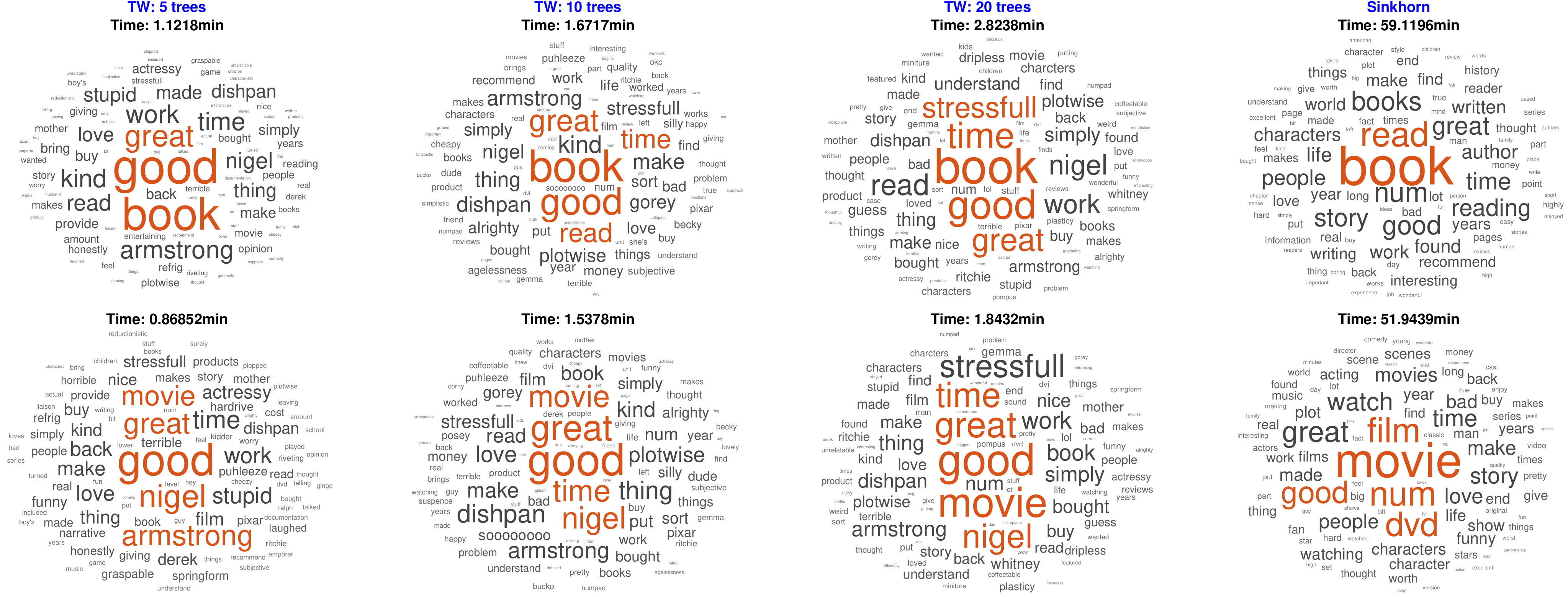}
  \end{center}
  \caption{An illustration for the word cloud (a visual representation of text data in which the higher corresponding weight a word has, the more prominently it is displayed) and time consumption of TW barycenter (with unconstrained free-support setting), and Sinkhorn-based barycenter (with fixed-support setting) for class ID1 (in the $1^{st}$ row) and class ID2 (in the $2^{nd}$ row) on AMAZON dataset. Moreover, TW barycenter was evaluated with $1$ CPU while Sinkhorn-based barycenter was run with multiple CPUs (40 2.80GHz-CPUs).}
  \label{fg:Amazon_TWBarycenter_Sinkhorn_ID12}
\end{figure*}

\subsection{Scalable Bayes}
\label{subsec:scale_Bayes}

We next study another application of TW barycenter and its variants to the scalable approximation of posterior distribution in Bayesian inference under the massive data settings. In particular, we assume that $Y_{i} \mid_{i \in [n]}$ are i.i.d. samples from the true probability distribution $P_{\theta^*}$ with density function $f(\cdot \mid \theta^*)$ where $\theta^* \in \Theta \subset \Rspace^{d}$ is a true parameter. Under the Bayesian framework, we endow the parameter $\theta$ with a prior distribution with its density $\pi(\theta)$. Then, given $\theta$, we fit $Y_{i} \mid_{i \in [n]}$ to be i.i.d. from $P_{\theta}$ with density function $f(\cdot \mid \theta)$. Given that setting, the posterior distribution of $\theta$ given the data $Y_{i} \mid_{i \in [n]}$ is given by
\begin{align}
    \Pi \left(\theta \mid \{Y_{i}\}_{\mid_{i \in [n]}} \right) : = \dfrac{\prod \limits_{i = 1}^{n} f(Y_{i} \mid \theta) \pi( \theta)}{\int \limits_{\Theta} \prod \limits_{i = 1}^{n} f(Y_{i} \mid \theta) \pi( \theta) \text{d} \theta}.
\end{align}
In general, the posterior distribution $\Pi (\theta \mid \{Y_{i}\}_{\mid_{i \in [n]}})$ is computationally intractable due to the expensive term in its denominator. Even though sampling methods, such as MCMC, are widely used to approximate the posterior, their computations are notoriously expensive when the sample size $n$ is large.

Recently, Srivastava et al.~\cite{Srivastava-2018-Scalable} proposed an efficient divide-and-conquer approach, which is termed as Wasserstein posterior (WASP), for approximating the posterior distribution based on the Wasserstein barycenter, which has favorable practical performance over other state-of-the-art methods. In the paper, we use this approach with TW barycenter and its variants, which refers to as \emph{tree-Wasserstein posterior}, and show that our model has a better running time than the previous method. 

The crux of divide-and-conquer approach is to divide the data $Y_{i} \mid_{i \in [n]}$ into $m$ machines where each machine has $k$ data, i.e., $n = m k$. To simplify the presentation, we denote $Y_{ij}$ as the data in $i$-th machine for $i \in [m]$ and $j \in [k]$. The subset posterior distribution given the data in machine $i$ is given by \footnote{Note that the stochastic approximation trick which raises the likelihood to the power of $m$ is used to ensure that subset posteriors and the full posterior have variances of the same order.}
\begin{align*}
    \Pi_{i} \left(\theta \mid \{Y_{ij}\}_{\mid_{j \in [k]}} \right) : = \dfrac{\biggr(\prod \limits_{j = 1}^{k} f(Y_{ij} \mid \theta)\biggr)^{m} \pi( \theta)}{\int \limits_{\Theta} \biggr(\prod \limits_{j = 1}^{k} f(Y_{ij} \mid \theta)\biggr)^{m} \pi( \theta) \text{d} \theta}.
\end{align*}
In general, the subset posterior distributions are still intractable to compute due to their expensive denominators. However, as $k$ is sufficiently small, we can use MCMC methods to approximate these posteriors. In particular, we assume that $\theta_{ij} \mid_{j \in [N]}$ are the samples drawing from $\Pi_{i} \left(\theta \mid \{Y_{ij}\}_{\mid_{j \in [k]}}\right)$ via MCMC methods. Then, we can approximate these posterior distributions by the empirical measures of their samples, which can be defined as 
\[
\widehat{\Pi}_{i} \left(\theta \mid \{Y_{ij}\}_{\mid_{j \in [k]}} \right) : = \frac{1}{N} \sum_{j = 1}^{N} \delta_{\theta_{ij}},
\]
for $i \in [m]$. The TW posterior (tree-WASP) combines the approximate posterior distributions $\widehat{\Pi}_{i} \left(\theta \mid \{Y_{ij}\}_{\mid_{j \in [k]}} \right)$ for $i \in [m]$ through their TW barycenter, which can be formulated below:
\begin{align}
    & \hspace{-0.7 em}\overline{\Pi} \left( \theta \mid \{Y_{i}\}_{\mid_{i \in [n]}} \right) : =  \nonumber \\
    &\mathop{\arg \min}\limits_{\Pi} \dfrac{1}{m} \sum_{i = 1}^{m} d_{\TW} \biggr(\Pi, \widehat{\Pi}_{i} \left(\theta \mid \{Y_{ij}\}_{\mid_{j \in [k]}} \right)\biggr). \label{eq:tree_Wasp}
\end{align}
The tree-WASP $\overline{\Pi} \left( \theta \mid \{Y_{i}\}_{\mid_{i \in [n]}}\right)$ serves as an approximation for the original posterior distribution $\Pi \left(\theta \mid \{Y_{i}\}_{\mid_{i \in [n]}} \right)$. Similarly, one can sample several random tree metrics and leverage the multiple-tree variants for TW barycenter to reduce the clustering sensitivity effect. The pseudo-code for computing the tree-WASP is presented in Algorithm~\ref{alg:tree_scalable_Bayes}.
\begin{algorithm}
   \caption{Tree-Wasserstein posterior}
   \label{alg:tree_scalable_Bayes}
\begin{algorithmic}
   \STATE {\bfseries Input:} Data $Y_{i} \mid_{i \in [n]}$ and the number of machines $m$.
   \STATE {\bfseries Output:} Tree-WASP $\overline{\Pi} \left( \theta \mid \{Y_{i}\}_{\mid_{i \in [n]}} \right)$.
   \FOR{\textbf{each} subset posterior $\Pi_{i} \left(\theta \mid \{Y_{ij}\}_{\mid_{j \in [k]}} \right)$}
	\STATE Draw samples $\theta_{ij} \mid_{j \in [N]}$. 
    \ENDFOR
    \STATE Compute $\overline{\Pi} \left( \theta \mid \{Y_{i}\}_{\mid_{i \in [n]}} \right)$ as TW barycenter from approximate subset posteriors $\widehat{\Pi}_{i} \left(\theta \mid \{Y_{ij}\}_{\mid_{j \in [k]}} \right)$ for $i \in [m]$ (Equation~\eqref{eq:tree_Wasp}).
\end{algorithmic}
\end{algorithm}

\begin{figure*}
\begin{centering}
\includegraphics[width=2\columnwidth]{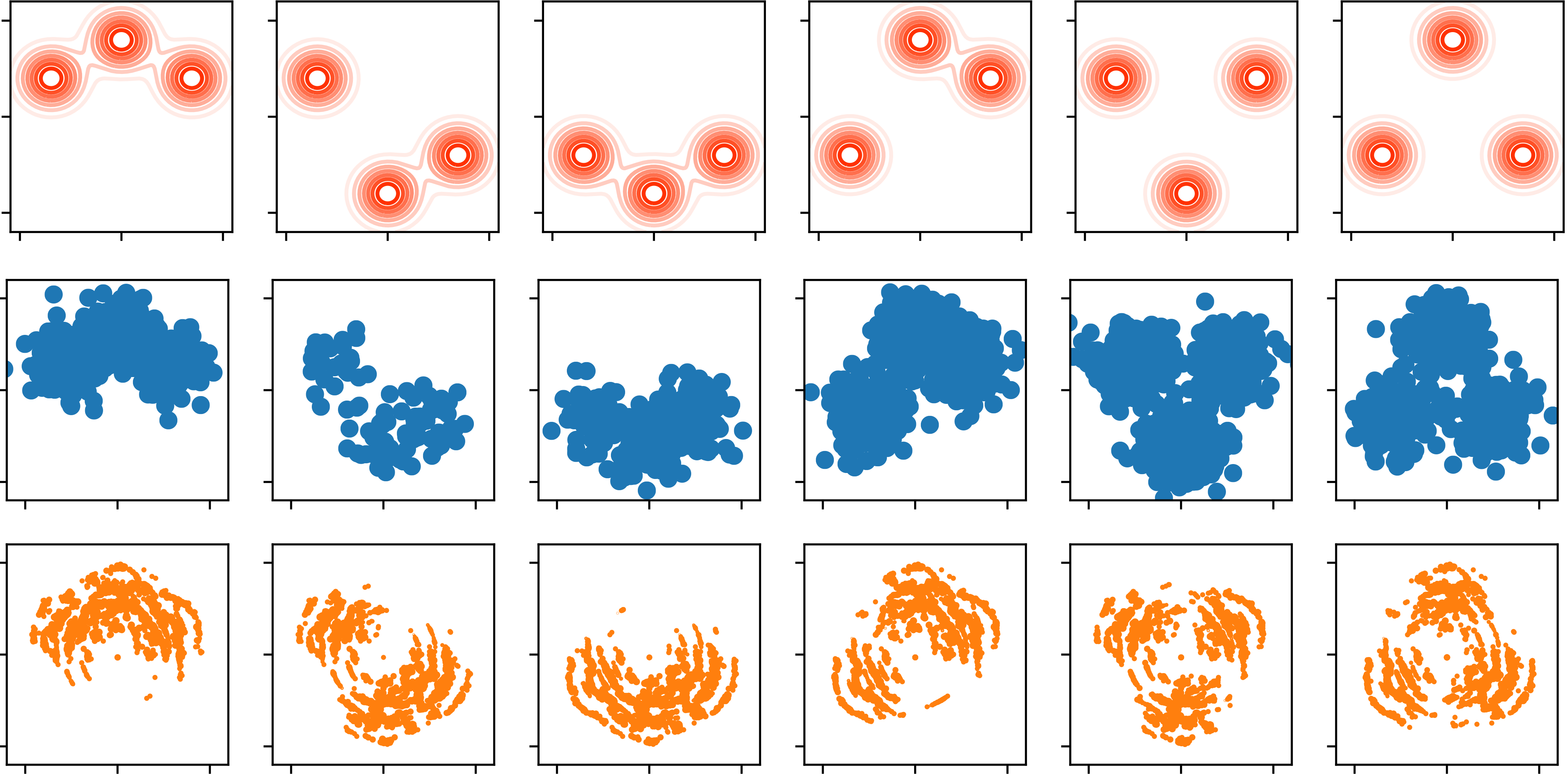}
\par\end{centering}
\caption{\label{fg:syn_GMM} An illustration of synthetic multilevel Gaussian data: (i) the ground truth Gaussian mixture of six clusters in the first row; (ii) the generated data from above mixtures which are used to fit tree-MWM model in the second row; (iii) the learned mixtures with our tree-MWM model from the the above generated data in the third row. Each data point in the last row is scatter with its corresponding weight of the learned barycenter.}
\end{figure*}

\section{Related work}\label{sec:related_work}

Tree metric has recently become an efficient tool to scale up a computation for optimal transport, i.e., tree-(sliced)-Wasserstein distance~\cite{le2019tree}, Gromov-Wasserstein~\cite{le2019fast} for large-scale applications. Note that, sliced-Wasserstein (SW) is a special case of tree-sliced-Wasserstein where a tree is a chain. Moreover, SW relies on one-dimensional projection for supports. Therefore, SW is fast, but limits its ability to capture a structure of high-dimensional distributions~\cite{pmlr-v97-liutkus19a} while TW can remedy this problem by leveraging distributions of supports to sample tree metric, e.g., clustering-based tree metric sampling~\cite{le2019tree}.

Our proposed TW barycenter is built upon the TW distance~\cite{le2019tree}. However, Le et al.~\cite{le2019tree} only derived the closed-form computation for TW distance, its tree-sliced variants, and their corresponding positive definite kernels. In this work, we further leveraged TW distance into a more complex parametric inference problem known as barycenter problem. We then proposed novel and efficient algorithms for TW barycenter and its variants with a bounded constraint on the number of supports by exploiting a tree mapping with simpler geometry (i.e., $\ell_1$ metric) for probability measures with OT geometry (see Algorithm~\ref{alg:TW_Barycenter}), and leveraging the \textit{center of mass} for a probability measure in a tree (see Algorithm~\ref{alg:Constrained_TW_Barycenter}).   

Additionally, Wasserstein barycenter is one of the main bottlenecks of MWM~\cite{Ho-ICML-2017} for multilevel clustering, and WASP~\cite{Srivastava-2018-Scalable} for scalable Bayes in large-scale applications. Both the proposed TW barycenter and SW barycenter~\cite{Bonneel_2015_sliced} can be applied to scale up a computation of multilevel clustering, and scalable Bayes. However, SW barycenter may suffer the high-dimensional curse as in SW due to its intrinsic usage of one-dimensional projections. Therefore, the proposed TW barycenter becomes a better candidate for multilevel clustering and scalable Bayes in large-scale applications. Moreover, the proposed Algorithm~\ref{alg:TW_Barycenter} for TW barycenter is very efficient since it can exploit a simpler geometry (i.e., $\ell_1$ metric for tree mapping) for a barycenter problem where one can not do the same trick for SW barycenter. Last but not least, recall that SW is a special case of TW~\cite{le2019tree}.

\section{Experiments}
\label{sec:experiments}


We first validate performances of TW barycenter by comparing it with Sinkhorn barycenter on AMAZON dataset in Section~\ref{sec:exp_documents}. Then, we carried out experiments with various large-scale datasets for multilevel clustering and scalable Bayes inference problems in Section~\ref{sec:exp_multilevel}, and Section~\ref{sec:exp_scal_bayes} respectively.

\subsection{Wasserstein barycenter problem for documents with word embedding }
\label{sec:exp_documents}

We evaluated TW barycenter on AMAZON, a textual dataset containing $4$ classes, and each class has $2000$ documents. We used the $word2vec$ word embedding \cite{mikolov2013distributed}, pre-trained on Google News\footnote{https://code.google.com/p/word2vec}. It includes about $3$ million words/phrases, mapped into $\RR^{300}$. We dropped all SMART stop words \cite{salton1988term}, and words which are not in the pre-trained $word2vec$ as in \cite{kusner2015word, le2019tree}. For TW barycenter, we optimize for both supports and its corresponding weights, i.e., free-support Wasserstein barycenter setting. We used clustering-based tree metric method \cite{le2019tree} to sample tree metrics where we used its suggested parameter, set $\kappa=4$ for the number of clusters for the farthest-point clustering, and $H_{\Tt}=6$ for the predefined deepest level of the constructed tree. For Sinkhorn barycenter, we optimize barycenter under fixed-support setting. We used Euclidean ground metric, set $500$ for the entropic regularization parameter, and did $100$ iterations (when we performed more iterations, e.g. $200$, Sinkhorn-based barycenter suffered a numerical problem)\footnote{For Sinkhorn-based barycenter with fixed-support setting, we followed https://github.com/gpeyre/2014-SISC-BregmanOT/tree/master/code/barycenters}.

Figure~\ref{fg:Amazon_TWBarycenter_Sinkhorn_ID12} illustrates the word cloud result, a visual representation of text data in which the higher corresponding weight a word has, the more prominently it is displayed, and time consumption of TW barycenter, and Sinkhorn barycenter for classes ID1 and ID2 on AMAZON dataset. The time computation of TW barycenter is much less than that of Sinkhorn barycenter. Moreover, note that Sinkhorn barycenter was run parallelly on a $40$-CPU cluster (Intel(R) Xeon(R) CPU E7-8891 v3 2.80GHz), and required at least about $50$GB RAM while TW barycenter was evaluated with a single CPU. Further experimental results for different parameters and results for classes ID3 and ID4 can be seen in the supplementary.

\subsection{Multilevel clustering problem}
\label{sec:exp_multilevel}

In this section, we demonstrate the efficiency of our proposed algorithms with a synthetic dataset. We define six clusters of data, each of which is a mixture of three 2-dimensional Gaussian components. The ground truth of six random data mixtures is shown in the first row of Figure \ref{fg:syn_GMM}.  We uniformly generated $100$ groups of data, each group belongs to one of the six aforementioned clusters. Once the clustering index of a data group was defined, we generated $100$ data points from the corresponding mixture of Gaussian distributions. The second row of Figure \ref{fg:syn_GMM} shows the distribution of data for each cluster. We ran the proposed tree-Wasserstein Multilevel Clustering (tree-MWM) algorithm (Algorithm~\ref{alg:tree_Wasserstein_multilevel}) with 50 trees of 5 deep levels, 4 branches for each node. The algorithm can approximately recover ground truth distributions of generated data as shown in the third row of Figure \ref{fg:syn_GMM}.

In the following experiment, we aim to demonstrate the efficiency of our proposed method for large-scale dataset settings. In order to have a sufficiently large number of samples, we use a synthetic dataset similar to the previous experiment except that the number of data dimensions is 100. However, we generated $50000$ data points from the corresponding mixture of Gaussian distributions. Given our simulated setting, the total number of data points in 100 groups is 5 million. 

\begin{table*}
 \caption{Running time and memory usage of our tree-MWM and MWM \cite{Ho-ICML-2017} for the multilevel clustering.}
\label{table:memory}
\begin{centering}
\begin{tabular}{ccccc|ccc}
\cline{2-8} 
 &  & \multicolumn{3}{c}{Running time (s)} & \multicolumn{3}{c}{Memory usage (GB)}\tabularnewline
\cline{2-8} 
 & K & 30 & 100 & 300 & 30 & 100 & 300\tabularnewline
\hline 
MWM~\cite{Ho-ICML-2017} &&12794& 19678 & Out of memory  &25.94    & 33.95
 &Out of memory \tabularnewline
\hline 
slice-MWM &&19127&19388&  20601& 28.3 & 28.3 & 28.4 \tabularnewline
\hline 
tree-MWM &&\textbf{10170}& \textbf{11478}&  \textbf{14840}& \textbf{23.2} &\textbf{ 23.2} & \textbf{23.2} \tabularnewline
\hline 
\end{tabular}
\par\end{centering}
\end{table*}

We ran the tree-MWM algorithm with 10 trees of 5 deep levels, 4 branches for each node. We compare the performance of our algorithm with that of two baseline methods: the multilevel Wasserstein means (MWM) algorithm~\cite{Ho-ICML-2017}, the Sinkhorn-based version, which was run with \textit{only} 10 iterations (called MWM), and the slice-based version of MVM in which we use the slice Wasserstein to compute barycenter (denoted slice-MWM). All algorithms were set with the maximum number of local clusters as $k = 30$ while the number of global clusters $K \in \{30, 100, 300\}$. We ran the  Sinkhorn-based algorithm with the regularization parameter as $10$ which was chosen in the set of  $\{0.1, 1, 10, 100\}$.  For the slice-based version we choose the number of projections as $50$ which is also chosen from the set of $\{10, 50, 100, 1000\}$.  The parameters of two baseline methods are chosen based on the smallest distance between the learned and groundtruth cluster means. When $K = 30$, the running times of our proposed method outperform its baseline methods. Moreover, tree-MWM is also more efficient in terms of memory usage as depicted in Table~\ref{table:memory}. The reason is that MWM need to compute and store the cost matrix  of size $n\times k$ for \textit{each} data group where $n$ is the number of data points of a group ($n = 50000$) and $k$ is the number of local clusters while tree-MWM does not need to store these matrices but trees for \textit{all} data groups. Note that for a large number of local clusters, e.g.  $k=300$, MWM can not handle since there is not enough memory\footnote{More than 55GB memory is required.} while tree-MWM is still robust.

\subsection{Scalable Bayes problem}
\label{sec:exp_scal_bayes}

In this section, we demonstrate the efficiency of TW barycenter when applied to the scalable Bayes problem. Different from the divide-and-conquer approach in~\cite{Srivastava-2018-Scalable}, we use TW barycenter instead of standard Wasserstein barycenter to combine the subset posteriors in different machines to approximate the full posterior distribution.

In order to illustrate the scalable performance of our approach, we consider the large-scale setting of linear mixed effects model (cf. \cite[\S4.3]{Srivastava-2018-Scalable}). Suppose that we have $n=6000$ data groups with the total number of observations $s=100.000$. The number of observations in each group $s_i$ are uniform among groups, i.e. $\sum_i s_i=s$. Letting $X_i \in \RR^{s_i \times 4}$ and $Z_i \in \RR^{s_i \times 3}$,and  $\yb_i \in \RR^{s_i}$ be the observed features in the fixed and random effects components, and the response for data group $i$, respectively. The generative process of the linear mixed effects model follows
\begin{align}
  \label{eq:lme}
  \yb_i \mid \betab, \ub_i, \tau^2 \sim \Ncal_{s_i}(X_i \betab + Z_i \ub_i, \tau^2 I_{n_i}),\\ \quad \ub_i \sim \Ncal_r(\zero, \Sigma), \quad (i = 1, \ldots, n),\nonumber
\end{align}
where $\tau$, $\betab$, and $\Sigma$ are the model parameters. The priors for these parameters are chosen as $1/\tau^2 \sim \text{Half-Cauchy(0,2.5)}$, $\betab\sim\Ncal(\zero,\tau^2 I_{3})$, and $\Sigma\sim\text{LKJCorr}(2)$ which is LKJ correlation distribution \cite{lewandowski2009generating}. We generated data using the generative process in Equation (\ref{eq:lme}) with true $\betab=[-2,2,-2,2]$, $\tau^2=1$, and  $\Sigma = \diag(\sqrt{1}, \ldots, \sqrt{3}) \cdot R \cdot \diag(\sqrt{1}, \ldots, \sqrt{3})$ where $R$ is a symmetric correlation matrix with diagonal of one and $R_{12}=-0.40$, $R_{13}=0.30$, and $R_{23}=0.001$. We divided generated data into $50$ partitions and used the no-u-turn sampler~\cite{hoffman2014no} to simulate $10.000$ samples of each sub-posterior.
\begin{figure}
  \begin{center}
    \includegraphics[width=\columnwidth]{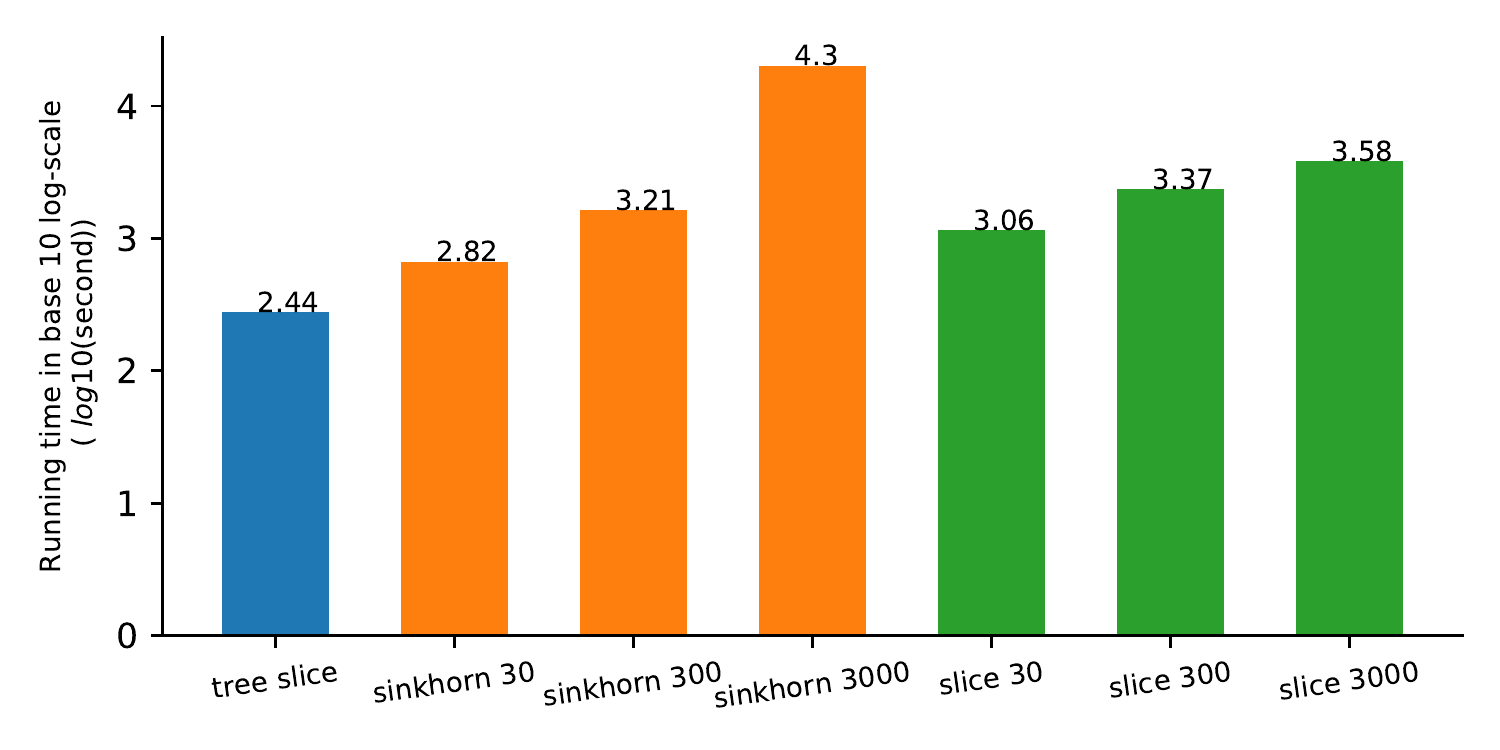}
  \end{center}
  \caption{Running time comparison between the proposed tree-Wasserstein barycenter and Sinkhorn-based \cite{Cuturi-2014-Fast} and slice-based \cite{Bonneel_2015_sliced} algorithms with various number of supports $k=30, 300, 3000$ for computing samples of posteriors of random covariance matrix $\Sigma$ for the scalable Bayes.}
  \label{fg:sb_runningtime}
\end{figure}
In the following experiments, we aim to estimate the random covariance matrix $\Sigma$ of $3\times 3$ dimensions. We then used the Sinkhorn-based algorithm \cite{Cuturi-2014-Fast}, slice-based Wasserstein barycenter\cite{Bonneel_2015_sliced}, and our proposed TW barycenter to estimate the posterior. We compared running time and the accuracy of both algorithms. For the Sinkhorn-based algorithm, we ran for \textit{only} 10 iterations with the best regularizer $\lambda=1$ while we used $100$ projection samples for slice-base algorithm. We use clustering-based 100 trees of 5 deep levels with $\kappa=4$ (the number of clusters for the farthest-point clustering in clustering-based tree metric construction procedure \cite{le2019tree}) for computing the barycenter. Figure \ref{fg:sb_runningtime} depicts the running time of our proposed algorithm and the Sinkhorn-based algorithm with different number of supports (samples) for estimating the posterior. Note that we do not need to specify the number of supports for our proposed algorithms which is bounded by the product of the number of nodes in tree and the number of trees. The TW barycenter algorithm requires less than half of the running time for the Sinkhorn-based algorithm with only 30 samples. 

\begin{figure}
  \begin{center}
    \includegraphics[width=\columnwidth]{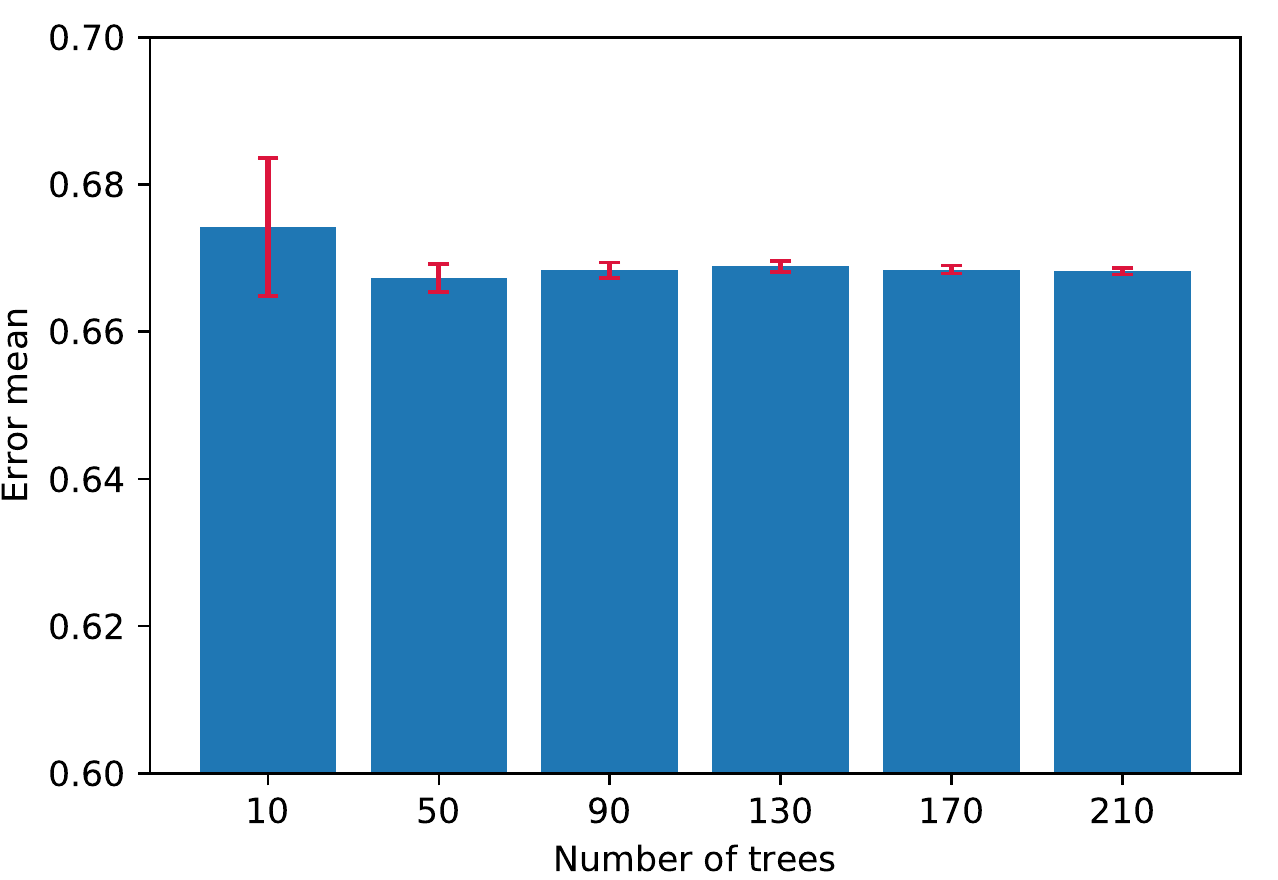}
  \end{center}
  \caption{Absolute error (and standard deviation) between the estimated and true covariance matrix $\Sigma$ for a different number of trees for the scalable Bayes.}
  \label{fg:error_means}
\end{figure}

We also investigate how the number of trees used in the proposed algorithms affects approximation performance. We estimate the covariance matrix $\Sigma$ using weighted samples from the barycenter and compare it with the true $\Sigma$. Figure \ref{fg:error_means} shows error means and standard deviations with a different number of trees. When the more number of trees is used, the more confidence we can obtain in the estimated parameters. Additionally, the error means between TW-based algorithm and two baseline methods, Sinkhorn-based and slice-base algorithms, depicted in Table \ref{tab:error_tree_vs_sinkhorn} show the efficiency of the proposed method.

\begin{table}[h]
 
\caption{\label{tab:error_tree_vs_sinkhorn}Absolute error between estimated and true covariance matrix of random effects for the scalable Bayes.}
\centering{}
\begin{tabular}{cll}
\cmidrule{3-3} 
&  & Error\tabularnewline
\midrule
{\small{}TW barycenter} & trees=10 & \textbf{0.67425 }\tabularnewline
\hline
\multirow{3}{*}{\shortstack{Sinkhorn \\ barycenter}} & k=30 & 0.751236 \tabularnewline
 & k=300 & 0.751239\tabularnewline
 & k=3000 & 0.751223\tabularnewline
 \hline
 \multirow{3}{*}{\shortstack{Sliced Wasserstein \\ barycenter}} & k=30 & 0.678854 \tabularnewline
 & k=300 & 0.846154\tabularnewline
 & k=3000 & 0.748187\tabularnewline
 \hline
\end{tabular}
\end{table}
\section{Conclusion}
\label{sec:discussion}
In this work, we propose efficient algorithm approaches for tree-Wasserstein barycenter and its variant (i.e., TW barycenter with a bounded constraint on the number of supports). By leveraging the favorable structure of tree metrics, we exploit tree mapping to solve the Wasserstein barycenter for probability measures on a simpler geometry (i.e., $\ell_1$ metric for tree mapping of probability measures) and further rely on some special tree properties (e.g., center of mass of a probability on a tree) to relax the variant of TW barycenter. Therefore, the proposed algorithms for TW barycenter and its variant are fast in computation. Consequently, based on the proposed TW barycenter, we scale up the multilevel clustering and scalable Bayes for large-scale applications. Empirically, we demonstrate the benefits of our algorithms against other baseline algorithms. We leave the question about efficient tree metric sampling for future work.

\section*{Acknowledgement}
We would like to thank Marco Cuturi for helpful discussion. TL acknowledges the support of JSPS KAKENHI Grant number 17K12745.

\balance
\bibliographystyle{plain}
\bibliography{Nhat}

\newpage
\onecolumn
\begin{center}
\textbf{\Large{Supplement to ``Tree-Wasserstein Barycenter \\ for Large-Scale Multilevel Clustering and Scalable Bayes"}}
\end{center}

In this supplementary material, we provide further experimental results for TW barycenter validation by comparing with Sinkhorn barycenter, multilevel clustering and scalable Bayes problems.

\appendix

\section{Further experimental results with TW barycenter versus Sinkhorn barycenter}
\label{sec:addition_exp}

\subsection{Further experiments with AMAZON dataset}

\begin{figure}[H]
  \begin{center}
    \includegraphics[width=\textwidth]{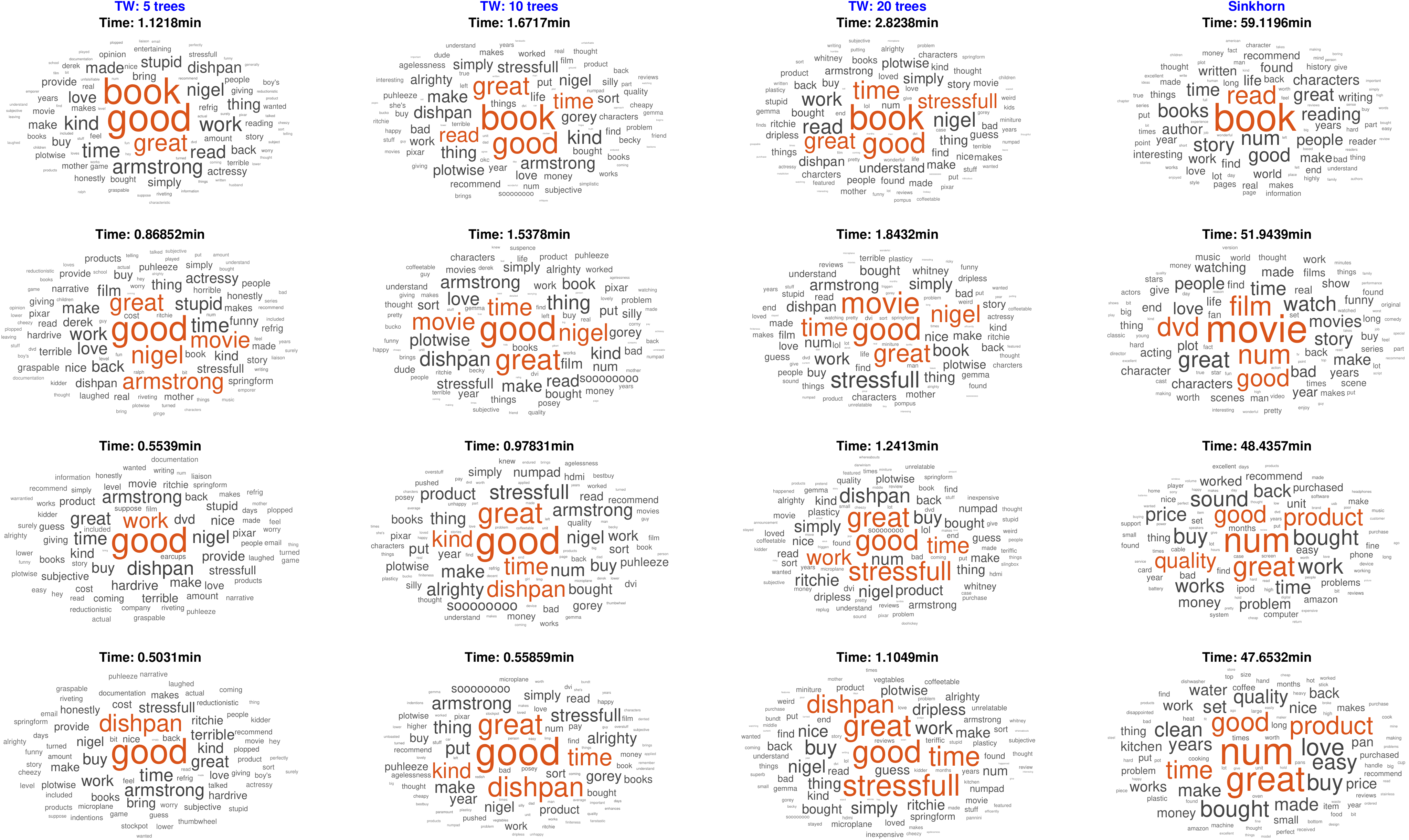}
  \end{center}
  \caption{An illustration for the word cloud (a visual representation of text data in which the higher corresponding weight a word has, the more prominently it is displayed) and time consumption of TW barycenter, and Sinkhorn barycenter on AMAZON dataset. Each row is corresponding to each class of documents.}
  \label{fg:Amazon_TWBarycenter_Sinkhorn_ALL}
\end{figure}

We show a comparison of word cloud---a visual representation of text data in which the higher corresponding weight a word has, the more prominently it is displayed---and time consumption of TW barycenter and Sinkhorn barycenter for each class on AMAZON dataset (each row is corresponding to each class: ID1, ID2, ID3, and ID4 respectively) in Figure~\ref{fg:Amazon_TWBarycenter_Sinkhorn_ALL}. 

\begin{figure}
  \begin{center}
    \includegraphics[width=\textwidth]{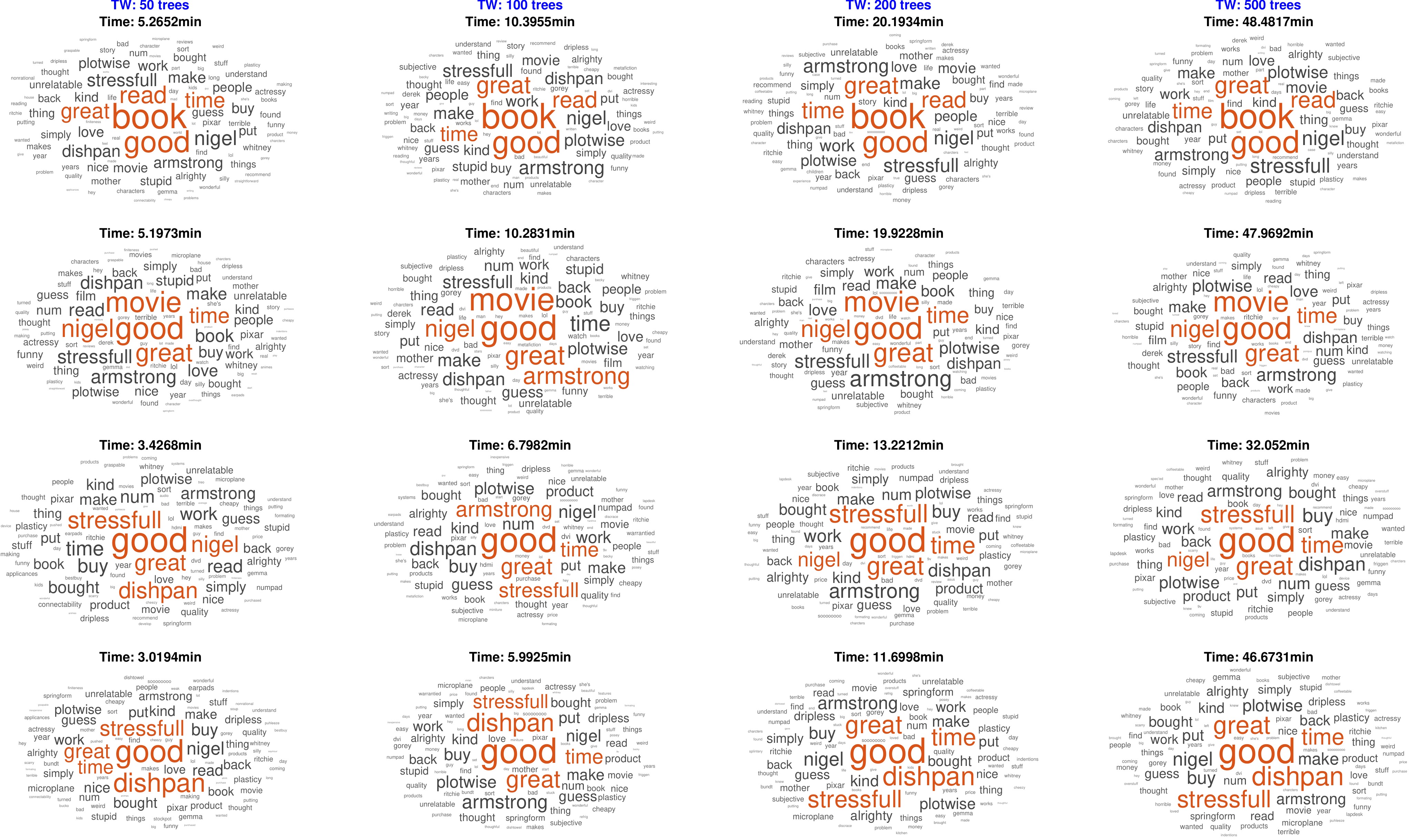}
  \end{center}
  \caption{An illustration for the trade-off between the quality and time consumption of TW barycenter for AMAZON dataset when one increases the number of trees. Each row is corresponding to each class of documents.}
  \label{fg:Amazon_TWBarycenter_ALL}
\end{figure}

Additionally, we illustrate further results for TW barycenter when one increases the number of tree metrics for each class on AMAZON dataset (each row is corresponding to each class: ID1, ID2, ID3, and ID4 respectively) in Figure~\ref{fg:Amazon_TWBarycenter_ALL}.

\begin{figure}
  \begin{center}
    \includegraphics[width=\textwidth]{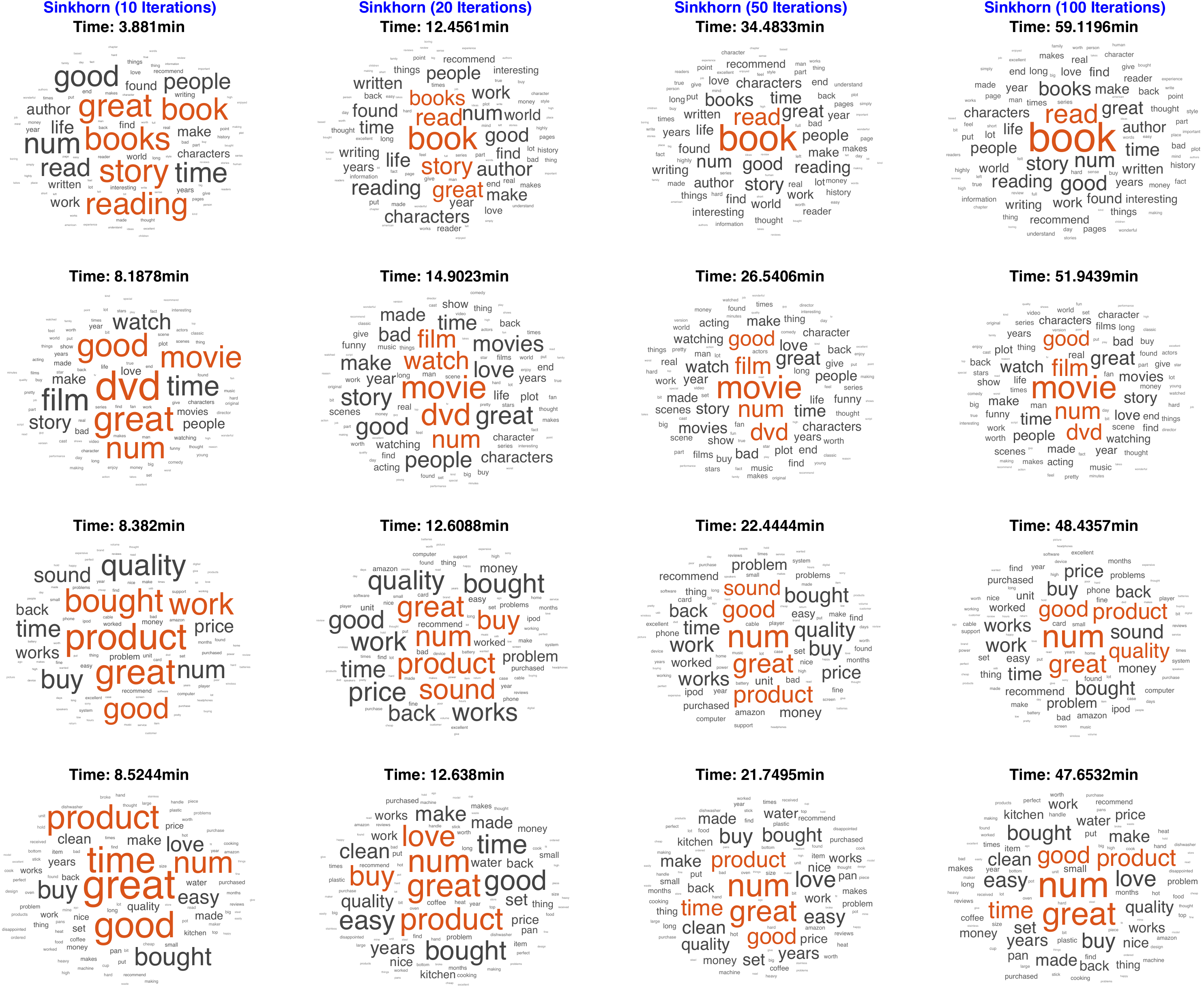}
  \end{center}
  \caption{An illustration for the trade-off between the quality and time consumption of Sinkhorn Wasserstein barycenter for AMAZON dataset when one uses more iterations (note that when the number of iterations is $200$, Sinkhorn barycenter suffered numerical problems (e.g. not a number problem). Each row is corresponding to each class of documents.}
  \label{fg:Amazon_SinkhornBarycenter_ALL}
\end{figure}

Moreover, we also illustrate further results for Sinkhorn barycenter when one increases the number of iterations for each class on AMAZON dataset (each row is corresponding to each class: ID1, ID2, ID3, and ID4 respectively) in Figure~\ref{fg:Amazon_SinkhornBarycenter_ALL}. Note that, when we increase more iterations (e.g. $200$ iterations), Sinkhorn barycenter suffered numerical problems (e.g. not a number problem).


\subsection{Wasserstein barycenter on MNIST handwritten digital images}
\label{sec:exp_mnist}

In this section, we illustrate the comparative performances of TW versus Sinkhorn for Wasserstein barycenters on the MNIST dataset. Note that, for Sinkhorn, each image is a $28 \times 28$ matrix of normalized intensity. So, each image can be regarded as an empirical measure where its supports are 2-dimensional location of pixels and its corresponding weights are the normalized intensities at those pixels. Then, we applied the fixed-support Sinkhorn barycenter (on the grid $28 \times 28$). Or, Sinkhorn barycenter only needs to optimize the corresponding weights for the $28 \times 28$ grid. While for TW, each image is represented as a point cloud of 2-dimensional positions of digit pixels, and ignore the positions of background pixels. After that, we used the unconstrained barycenter which optimizes both supports and corresponding weights.

\begin{figure}
  \begin{center}
    \includegraphics[width=0.6\textwidth]{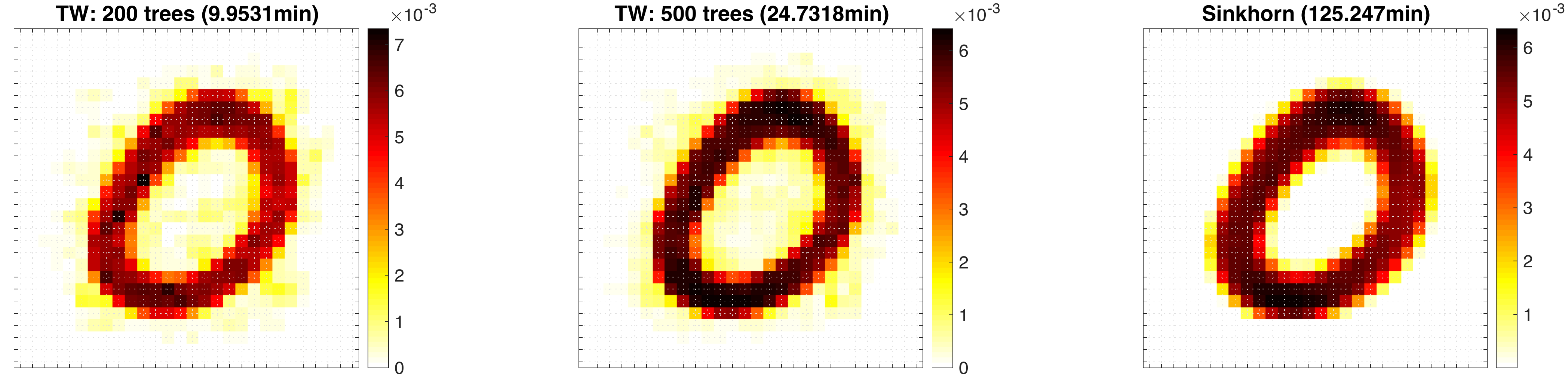}
  \end{center}
  \caption{A comparision between TW (free-support) and Sinkhorn (fixed-support) for Wasserstein barycenters with randomly $6000$ handwritten digital images of number $0$ on MNIST dataset.}
  \label{fg:MNIST_Num0}
\end{figure}

\begin{figure}
  \begin{center}
    \includegraphics[width=0.6\textwidth]{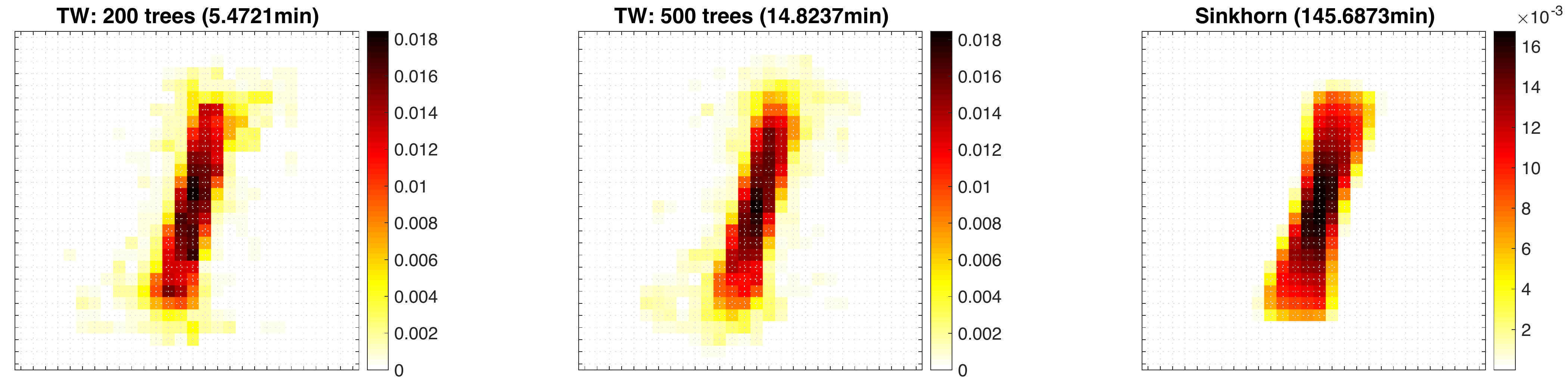}
  \end{center}
  \vspace{-12pt}
  \caption{A comparision between TW (free-support) and Sinkhorn (fixed-support) for Wasserstein barycenters with randomly $6000$ handwritten digital images of number $1$ on MNIST dataset.}
  \label{fg:MNIST_Num1}
  \vspace{-8pt}
\end{figure}

\begin{figure}
  \begin{center}
     \includegraphics[width=0.6\textwidth]{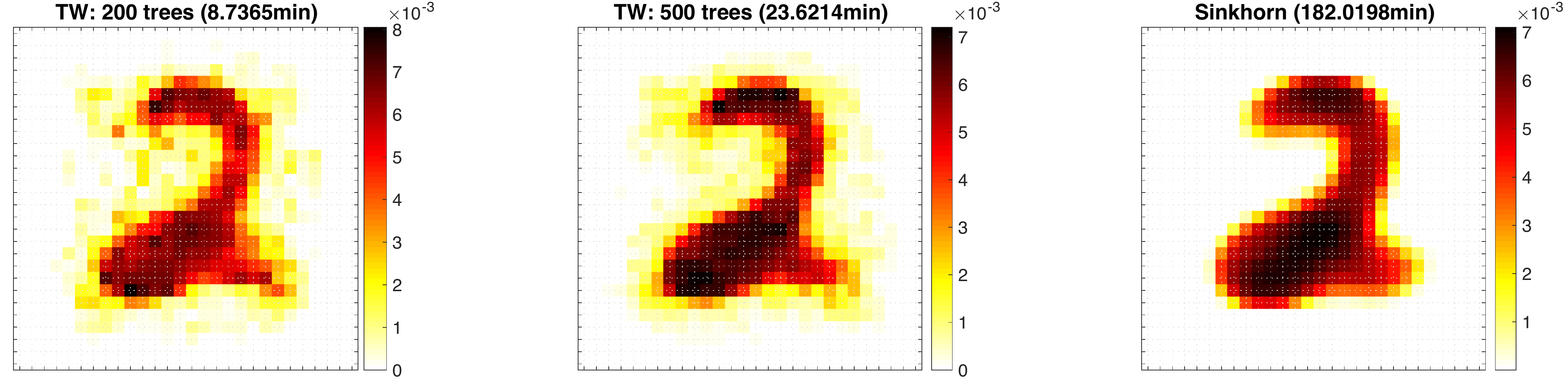}
  \end{center}
  \vspace{-12pt}
  \caption{A comparision between TW (free-support) and Sinkhorn (fixed-support) for Wasserstein barycenters with randomly $6000$ handwritten digital images of number $2$ on MNIST dataset.}
  \label{fg:MNIST_Num2}
  \vspace{-8pt}
\end{figure}

\begin{figure}
  \begin{center}
    \includegraphics[width=0.6\textwidth]{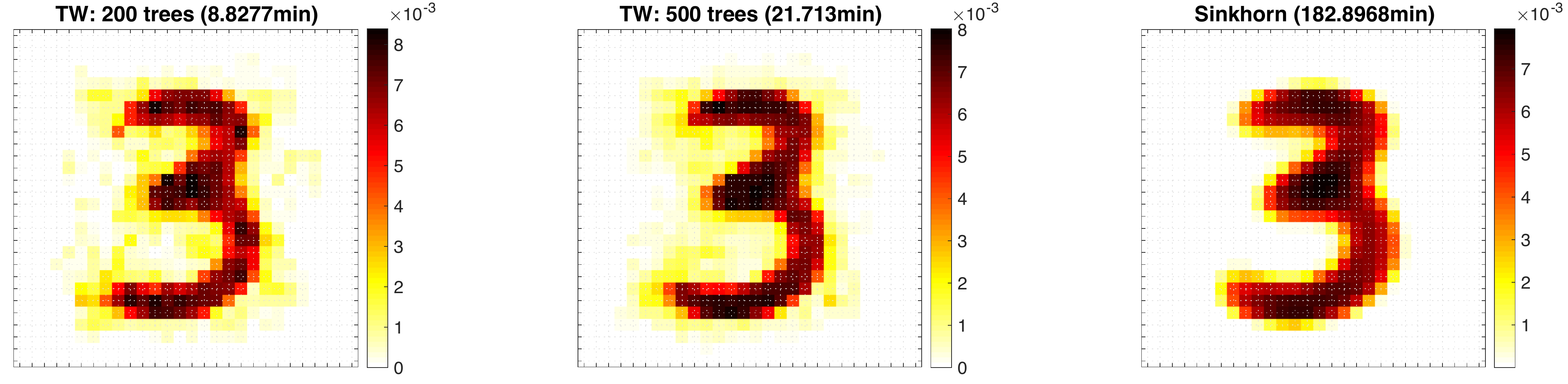}
  \end{center}
  \vspace{-12pt}
   \caption{A comparision between TW (free-support) and Sinkhorn (fixed-support) for Wasserstein barycenters with randomly $6000$ handwritten digital images of number $3$ on MNIST dataset.}
 \label{fg:MNIST_Num3}
  \vspace{-8pt}
\end{figure}

\begin{figure}
  \begin{center}
    \includegraphics[width=0.6\textwidth]{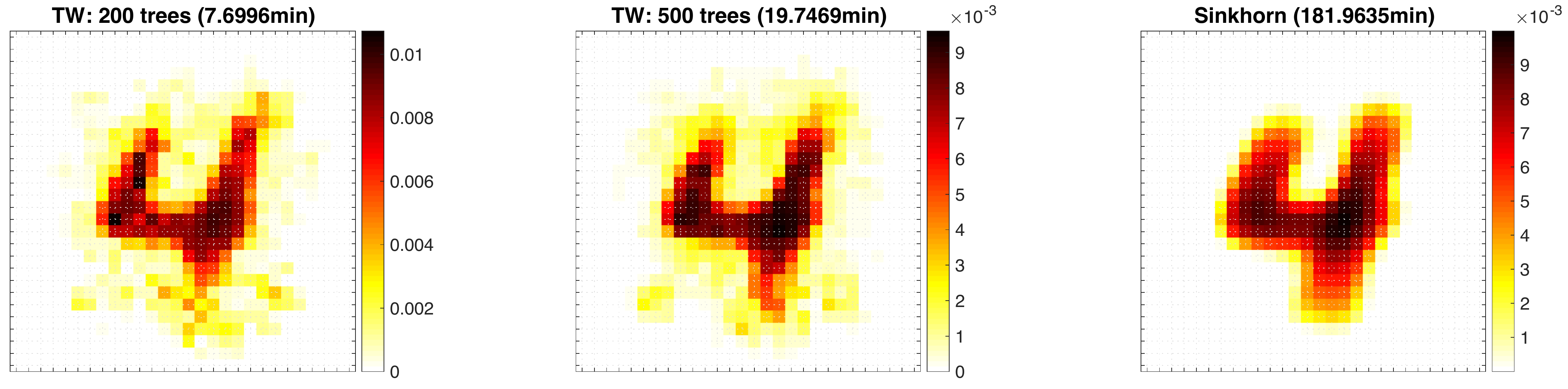}
  \end{center}
  \vspace{-12pt}
  \caption{A comparision between TW (free-support) and Sinkhorn (fixed-support) for Wasserstein barycenters with randomly $6000$ handwritten digital images of number $4$ on MNIST dataset.}
  \label{fg:MNIST_Num4}
  \vspace{-8pt}
\end{figure}

\begin{figure}
  \begin{center}
    \includegraphics[width=0.6\textwidth]{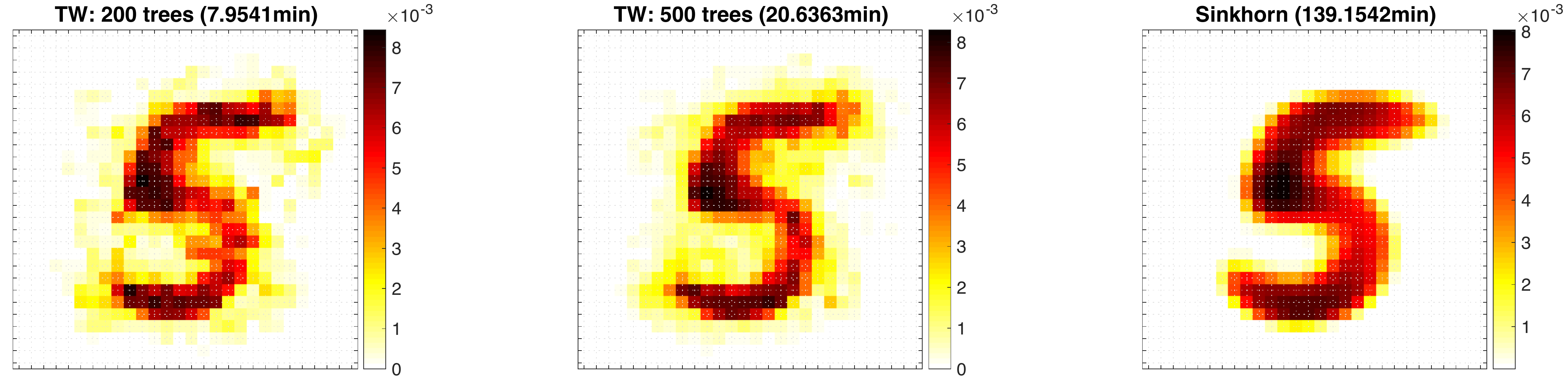}
  \end{center}
  \vspace{-12pt}
  \caption{A comparision between TW (free-support) and Sinkhorn (fixed-support) for Wasserstein barycenters with randomly $6000$ handwritten digital images of number $5$ on MNIST dataset.}
  \label{fg:MNIST_Num5}
  \vspace{-8pt}
\end{figure}

\begin{figure}
  \begin{center}
      \includegraphics[width=0.6\textwidth]{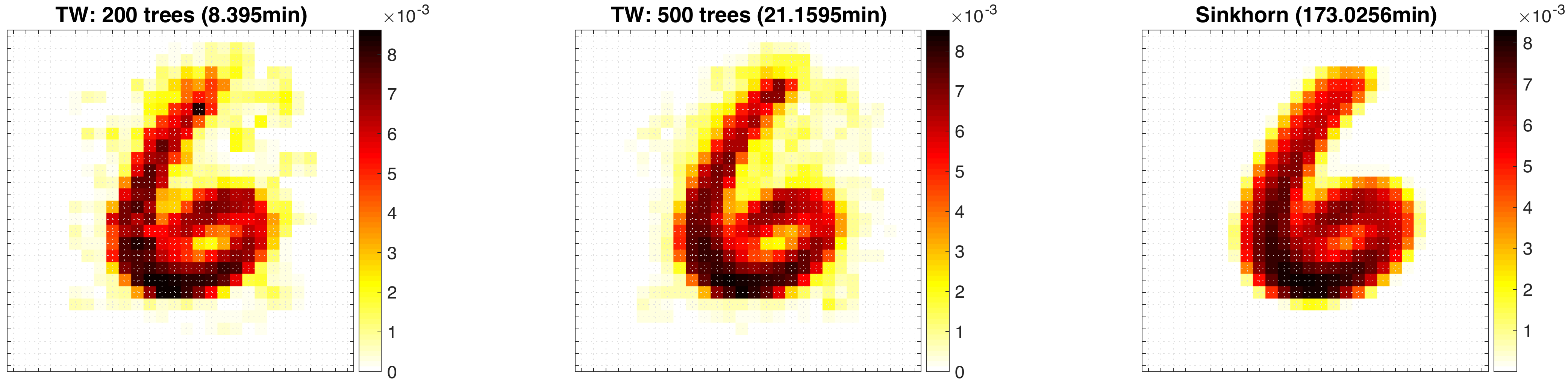}
  \end{center}
  \vspace{-12pt}
   \caption{A comparision between TW (free-support) and Sinkhorn (fixed-support) for Wasserstein barycenters with randomly $6000$ handwritten digital images of number $6$ on MNIST dataset.}
 \label{fg:MNIST_Num6}
  \vspace{-8pt}
\end{figure}

\begin{figure}
  \begin{center}
    \includegraphics[width=0.6\textwidth]{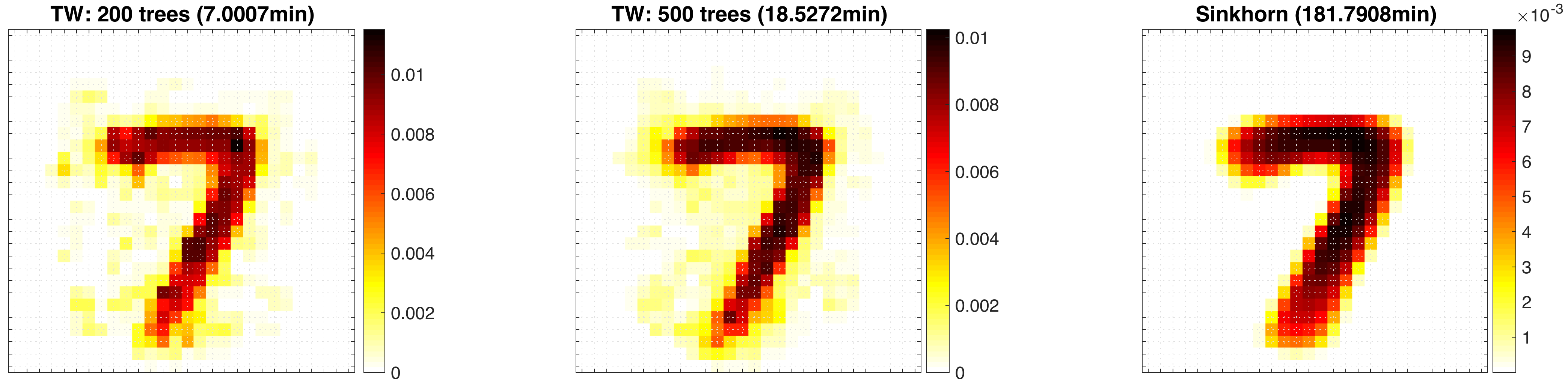}
  \end{center}
  \vspace{-12pt}
  \caption{A comparision between TW (free-support) and Sinkhorn (fixed-support) for Wasserstein barycenters with randomly $6000$ handwritten digital images of number $7$ on MNIST dataset.}
  \label{fg:MNIST_Num7}
 \vspace{-8pt}
\end{figure}

\begin{figure}
  \begin{center}
    \includegraphics[width=0.6\textwidth]{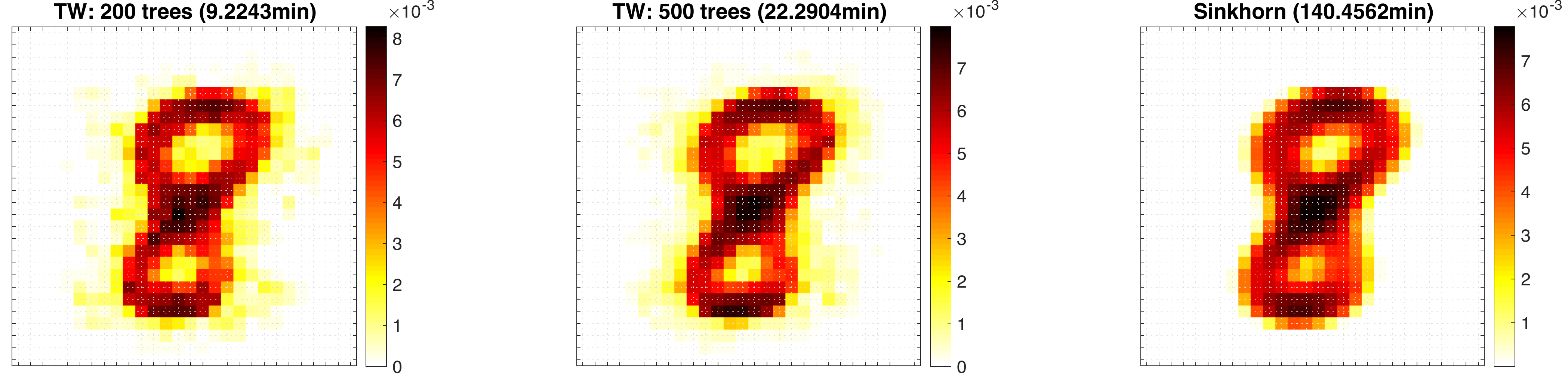}
  \end{center}
  \vspace{-12pt}
   \caption{A comparision between TW (free-support) and Sinkhorn (fixed-support) for Wasserstein barycenters with randomly $6000$ handwritten digital images of number $8$ on MNIST dataset.}
  \label{fg:MNIST_Num8}
  \vspace{-8pt}
\end{figure}

\begin{figure}
  \begin{center}
    \includegraphics[width=0.6\textwidth]{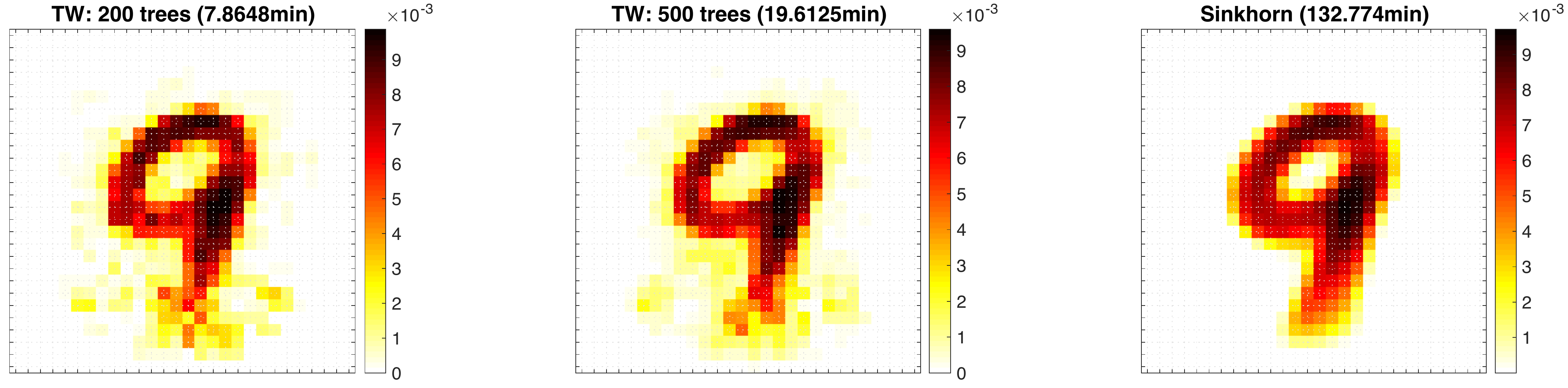}
  \end{center}
  \vspace{-12pt}
   \caption{A comparision between TW (free-support) and Sinkhorn (fixed-support) for Wasserstein barycenters with randomly $6000$ handwritten digital images of number $9$ on MNIST dataset.}
  \label{fg:MNIST_Num9}
  \vspace{-8pt}
\end{figure}

\begin{figure*}
  \begin{center}
    \includegraphics[width=0.75\textwidth]{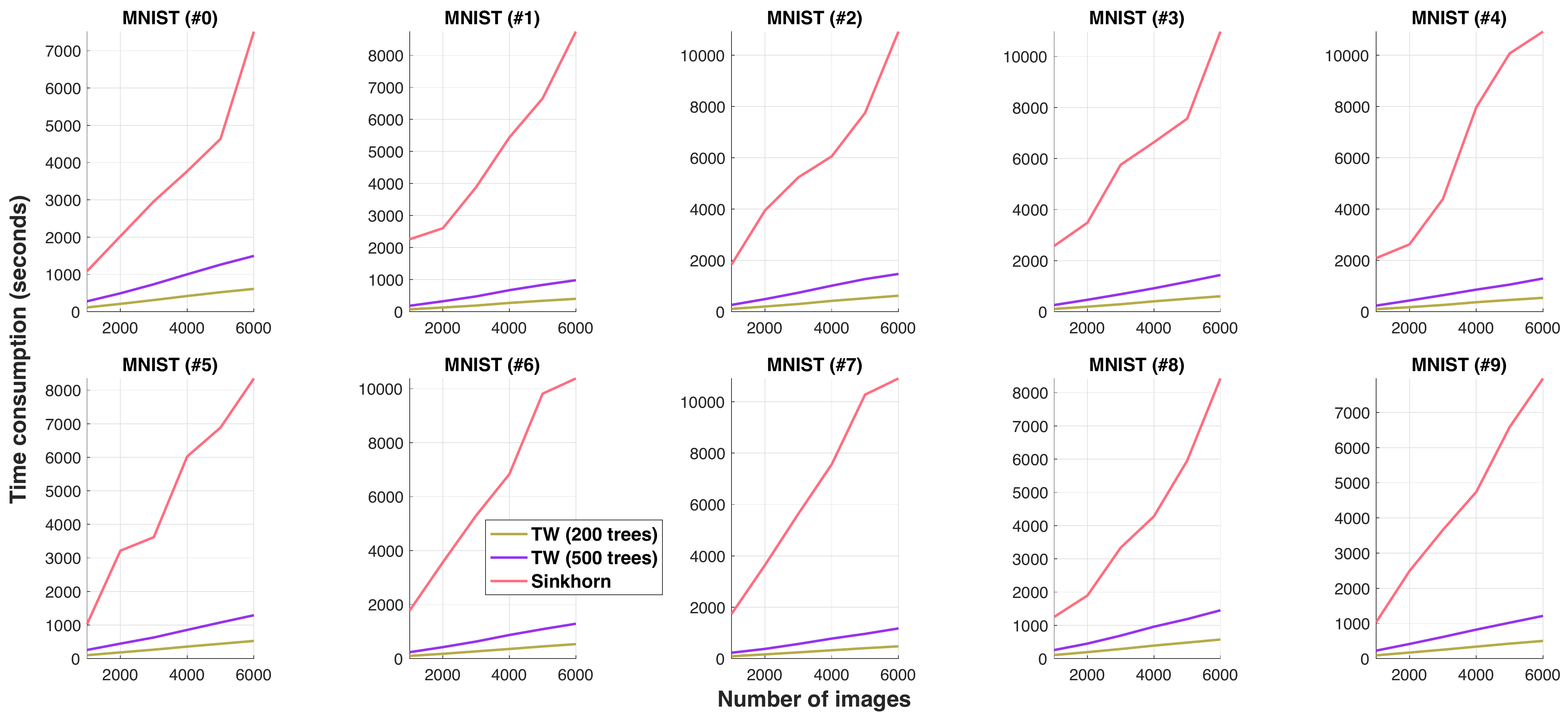}
  \end{center}
  \vspace{-12pt}
  \caption{A comparision of time consumption between TW (free-support) and Sinkhorn (fixed support) for Wasserstein barycenters with different numbers of handwritten digital images on MNIST dataset.}
  \label{fg:MNIST_AllNum_Time}
  \vspace{-8pt}
\end{figure*}

Figures~\ref{fg:MNIST_Num0}-\ref{fg:MNIST_Num9} illustrate a comparison between TW and Sinkhorn on Wasserstein barycenters where we randomly sample $6000$ images of each number $0-9$ on MNIST dataset. For Sinkhorn, we set $500$ for the entropic regulation parameter and use Euclidean distance as its ground metric. The barycenter of the Sinkhorn is optimized over a fixed grid $28 \times 28$, so it can be easily visualized by colormap (as pixel intensity). For TW, the number of trees is set to $200$ and $500$, and tree metrics are constructed by using the farthest-point clustering as in \cite{le2019tree} where we set $4$ for the number of clusters in the farthest-point clustering, and $10$ as the deepest level for the constructed tree. Figures~\ref{fg:MNIST_Num0}-\ref{fg:MNIST_Num9} show that the quality of TW barycenter is a trade-off with its computation when we increase the number of trees. Recall that, the TW barycenter optimizes both supports and corresponding weights. Therefore, for visualization, we employing a 4-neighbor interpolation to round the barycenter on the integer grid $28 \times 28$, then use colormap (as pixel intensity) as for Sinkhorn barycenter. We also illustrate a comparison of the computational time between TW and Sinkhorn for Wasserstein barycenters with different numbers of handwritten digital images on MINIST dataset in Figure~\ref{fg:MNIST_AllNum_Time}. Wasserstein barycenter with TW is faster than that of Sinkhorn, especially when we increase the number of images.

\section{Tree-Wasserstein multilevel clustering objective function}
\begin{figure}
  \begin{center}
    \includegraphics[width=0.75\textwidth]{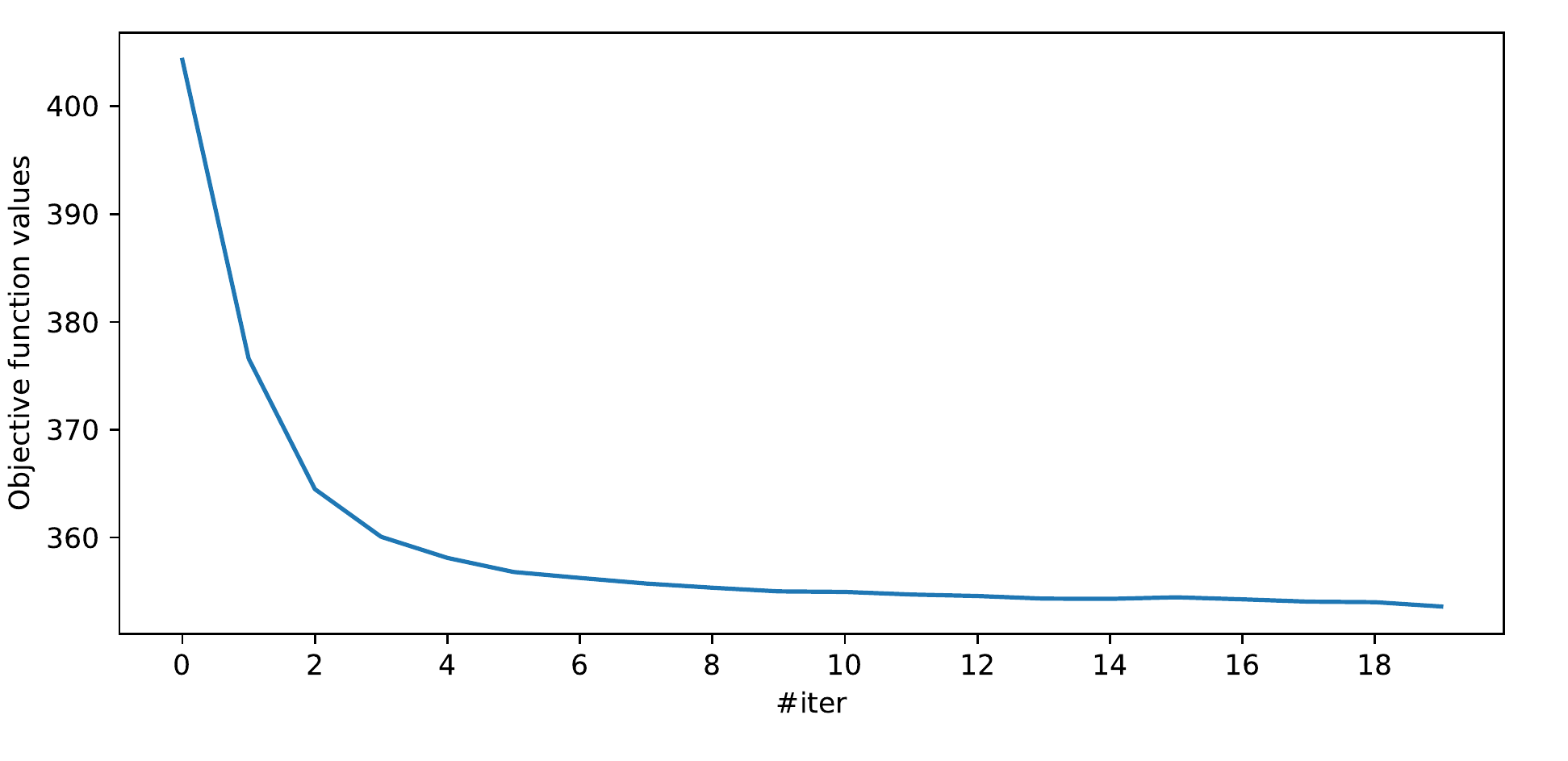}
  \end{center}
  \vspace{-12pt}
  \caption{An illustration of the evolution of the objective function values of tree-MWM (Algorithm~\ref{alg:tree_Wasserstein_multilevel}) over iterations.}
  \label{fg:TSMWM_obj_vals}
  \vspace{-8pt}
\end{figure}

In this experiment, we will experimentally demonstrate the convergence of Algorithm \ref{alg:tree_Wasserstein_multilevel} which uses the constrained tree-Wasserstein barycenter in Algorithm \ref{alg:Constrained_TW_Barycenter} as a sub-routine. Figure \ref{fg:TSMWM_obj_vals} depicts the convergence of tree-MWM algorithm in terms of objective function being decreased over iterations.

\section{Hyper-parameter tuning for two baseline methods using clustering performances}
\begin{figure}
  \begin{center}
    \includegraphics[width=0.75\textwidth]{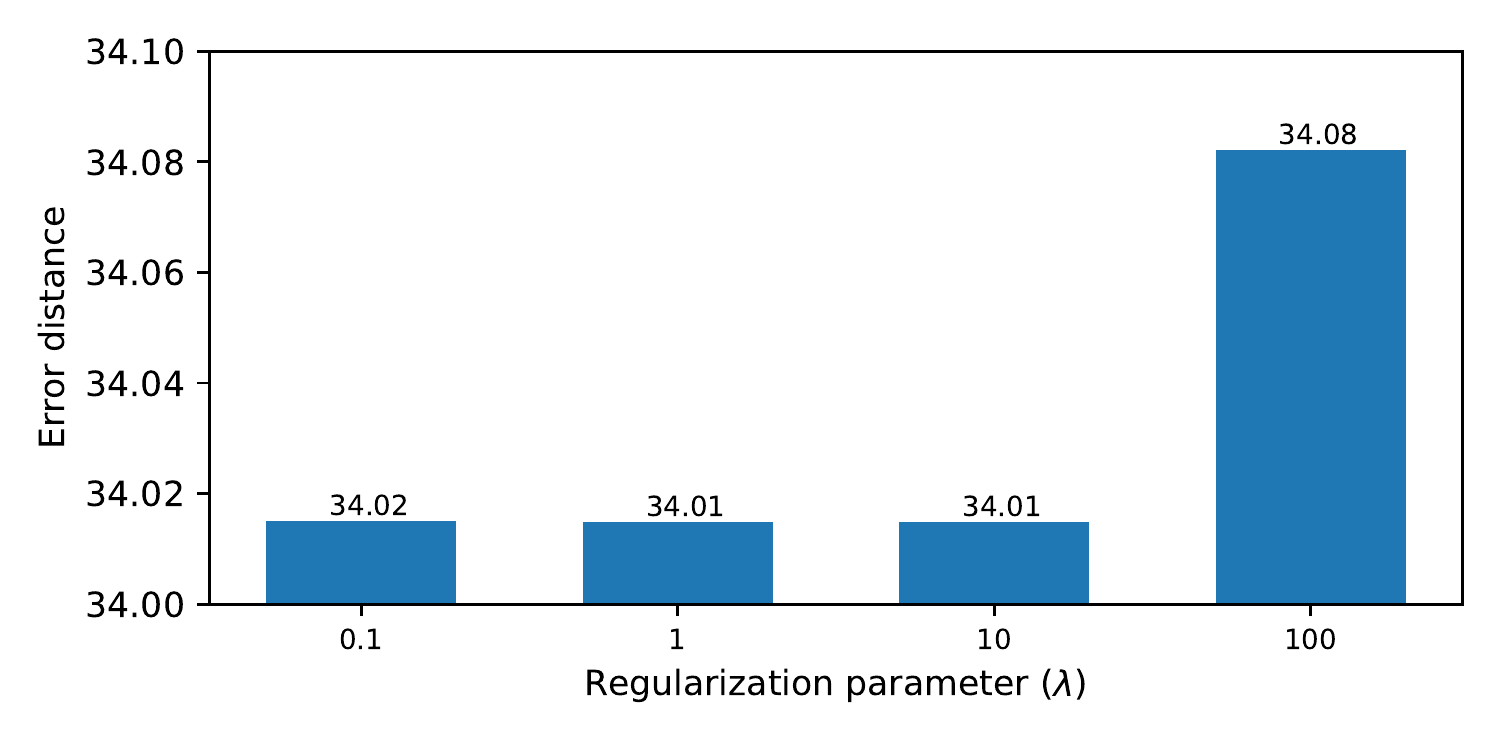}
  \end{center}
  \vspace{-12pt}
  \caption{Euclidean distance errors with respect to the regularization parameters for MVM algorithm.}
  \label{fg:MWM_vr_regularizers}
  \vspace{-8pt}
\end{figure}

\begin{figure}
  \begin{center}
    \includegraphics[width=0.75\textwidth]{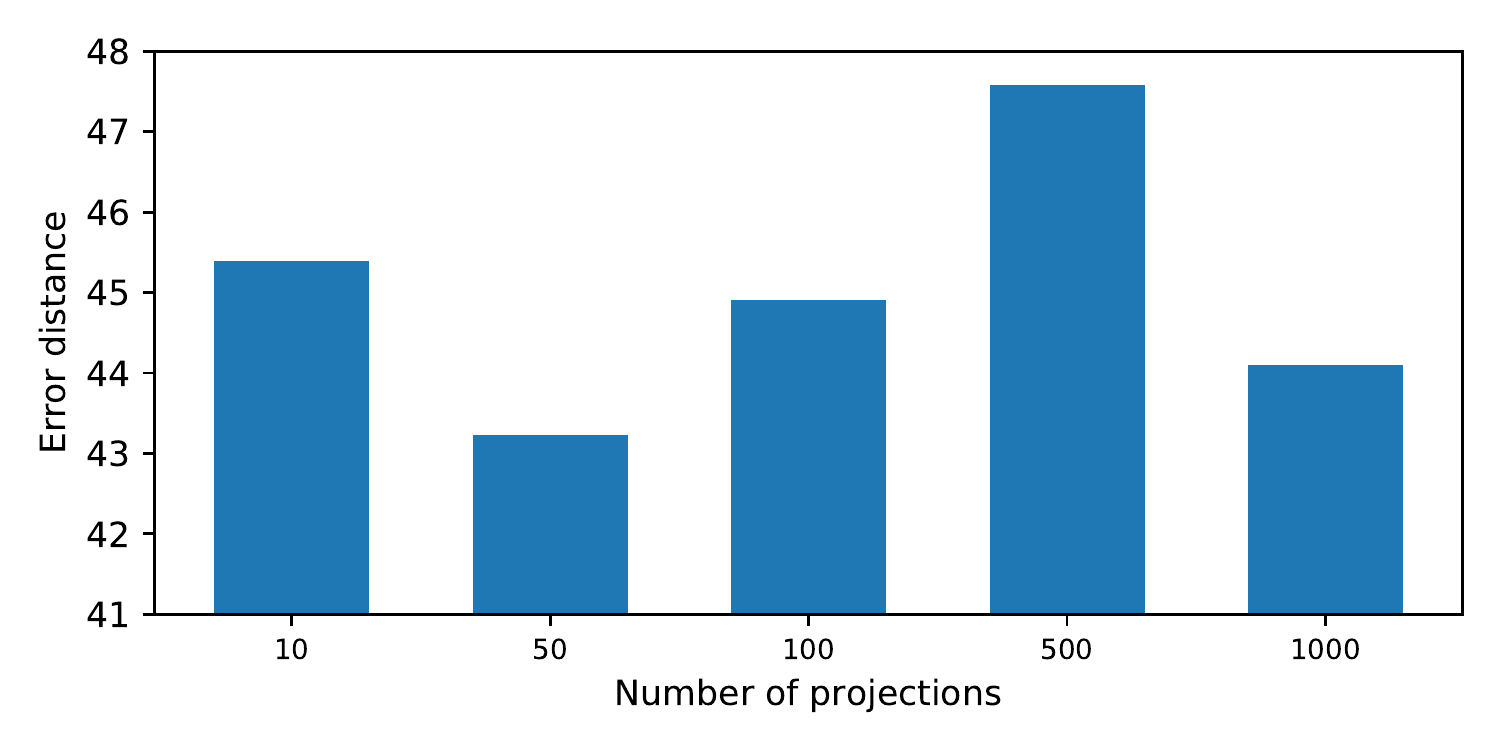}
  \end{center}
  \vspace{-12pt}
  \caption{Euclidean distance errors with respect to the numbers of projections for slice-MVM algorithm.}
  \label{fg:SMWM_vr_projections}
  \vspace{-8pt}
\end{figure}

In these experiments, we run experiments with different regularization parameters for MVM and different number of projection samples for slice-MVM. Figures \ref{fg:MWM_vr_regularizers} and \ref{fg:SMWM_vr_projections} depict the Euclidean distance error between the groundtruth and the learned means with respect to the regularization parameters and the number of projections.

\end{document}